\documentclass[a4paper,conference]{IEEEtran}
\usepackage{times}
\usepackage{epsfig}
\usepackage{graphicx}
\usepackage{amsmath}
\usepackage{amssymb}

\usepackage{algorithm}
\usepackage{algorithmicx}
\usepackage{algpseudocode}
\usepackage{subfigure}

\ifCLASSINFOpdf
\else
\fi
\begin{document}
%
\title{LDRNet: Large Deformation Registration Model\\ for Chest CT Registration}


\makeatletter
\newcommand{\linebreakand}{%
  \end{@IEEEauthorhalign}
  \hfill\mbox{}\par
  \mbox{}\hfill\begin{@IEEEauthorhalign}
}
\makeatother

\author{\IEEEauthorblockN{Cheng Wang}
\IEEEauthorblockA{Deepwise AI Laboratory\\
Beijing, China\\
Email: wangcheng@deepwise.com}
\and
\IEEEauthorblockN{Qiyu Gao}
\IEEEauthorblockA{Deepwise AI Laboratory\\
Beijing, China\\
Email: gaoqiyu@deepwise.com}
\and
\IEEEauthorblockN{Fandong Zhang}
\IEEEauthorblockA{Deepwise AI Laboratory\\
Beijing, China\\
Email: zhangfandong@deepwise.com}
\linebreakand
\IEEEauthorblockN{Shu Zhang}
\IEEEauthorblockA{Deepwise AI Laboratory\\
Beijing, China\\
Email: zhangshu@deepwise.com}
\and
\IEEEauthorblockN{Yizhou Yu}
\IEEEauthorblockA{Deepwise AI Laboratory\\
Beijing, China\\
Email: yuyizhou@deepwise.com}
}


%


\maketitle

\begin{abstract}
    Most of the deep learning based medical image registration algorithms focus on brain image registration tasks.
    Compared with brain registration, the chest CT registration has larger deformation,
    more complex background and region overlap.
    In this paper, we propose a fast unsupervised deep learning method, LDRNet,
    for large deformation image registration of chest CT images.
    We first predict a coarse resolution registration field, then refine it from coarse to fine.
    We propose two innovative technical components: 1) a refine block that is used to refine the registration field in different resolutions,
    2) a rigid block that is used to learn transformation matrix from high-level features.
    We train and evaluate our model on the private dataset and public dataset SegTHOR.
    We compare our performance with state-of-the-art traditional registration methods
    as well as deep learning registration models VoxelMorph, RCN, and LapIRN.
    The results demonstrate that our model achieves state-of-the-art performance for large deformation images registration and is much faster.
\end{abstract}


%
\IEEEpeerreviewmaketitle

\let\thefootnote\relax\footnote{This research was supported by the National Key Research and Development Program (No. 2019YFB1404804, No. 2019YFC0118101).}
\addtocounter{footnote}{-1}\let\thefootnote\svthefootnote

\section{Introduction}
Image registration is a fundamental technology used in many medical tasks,
such as surgical navigation, radiation therapy, and so on.
Rigid registration calculates rotation matrix and translation matrix before the whole image is transformed.
It has natural limitations in dealing with local deformation.
Nonrigid registration algorithms can be divided into traditional and deep learning methods.
Traditional algorithms often solve an optimization task by iterative methods,
which have disadvantages of computation and time cost.


\begin{figure}[h]
\centering
\includegraphics[width=0.7\linewidth]{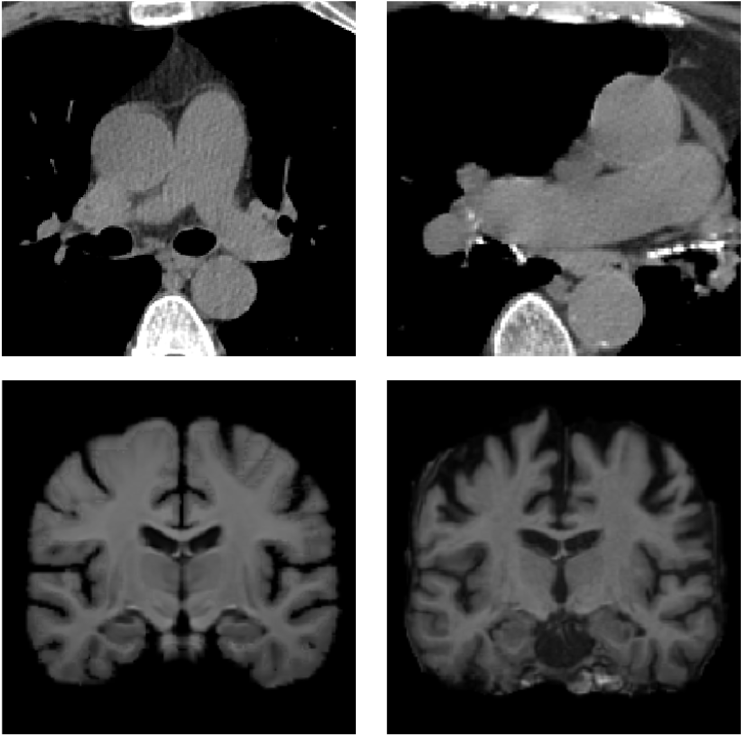}
\caption{Difference between large deformation chest images (up) and small deformation brain images (down)}
\label{fig:deformation}
\end{figure}

In recent years, many methods use deep learning to handle non-rigid registration tasks.
Some supervised deep learning methods are proposed.
Those methods require ground truth of the registration field, which is expensive and hard to obtain.
STN\cite{stn} proposes a method that achieves the nonrigid registration process by sampling from the moving image.
Since then many researchers focus on the unsupervised deep learning algorithms,
and VoxelMorph\cite{VoxelMorph1, VoxelMorph2, VoxelMorph3, VoxelMorph4} is one of the most commonly used methods.
The VoxelMorph formulates registration as a function that maps an input image pair to a deformation field.
However most unsupervised registration algorithms are designed for brain image registration,
which has a smaller deformation, cleaner background, and less region overlap compared with chest CT registration (Figure \ref{fig:deformation}).
They cannot achieve satisfactory registration results in the presents of large displacement between input images.

\begin{figure*}[h]
\includegraphics[width=0.8\linewidth]{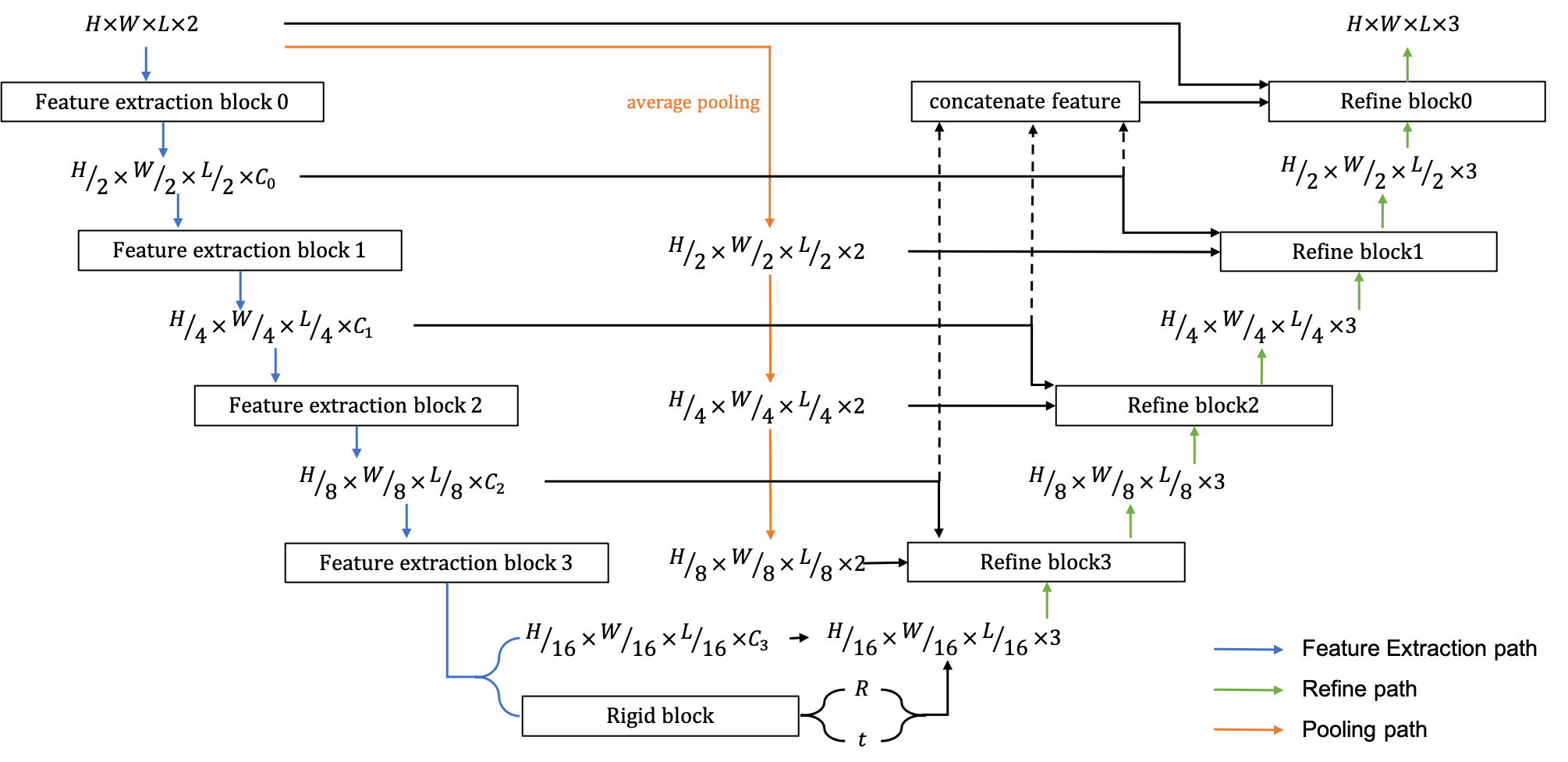}
\centering
\caption{Overview of our proposed registration model.
It includes a feature extraction path(blue), a refine path(green), a pooling path(orange) and a rigid block. The number of channels $C_0, C_1, C_2, C_3$ in the feature extraction path are 8, 16, 32, 64 respectively}
\label{fig:network}
\end{figure*}

We propose a novel coarse to fine registration network which can achieve registration between large deformation images.
The network refines the registration field from coarse resolution to fine resolution.
We design two blocks to solve the large deformation registration problems.
One is refine block which combines the fixed image, moving image, feature, and registration field to predict a refined registration field.
The other is rigid block which predicts transformation matrix using high-level features to provide a global matching.
We test our method on the private chest CT dataset and public dataset SegTHOR\cite{SegTHOR} separately.
The results are compared with state-of-art traditional algorithm deedsBCV, Ants, and deep learning algorithm VoxelMorph,
which demonstrate that our network has better accuracy and a faster speed.

In summary, the contributions of our work are:
\begin{itemize}
\item We propose a novel coarse-to-fine unsupervised deep learning registration model, including the refine block used to refine the registration field in different resolution.
\item Our method demonstrates computing speed advantage and the ability to deal with large-scale deformation problems on private and public chest CT datasets.
\end{itemize}


\section{Related Works}
\subsection{Traditional Registration Methods}
The traditional registration algorithms are well studied for years and many classic methods are proposed.
Elastic matching methods\cite{elastic1, elastic2} introduce the idea of multi-resolution.
Iterative closet point papers\cite{ICP1,ICP2,ICP3} propose a widely used natural image registration method, which is often used for point cloud registration.
In addition, there are symmetric normalize\cite{syn}, statistical parametric mapping\cite{statistic} and etc.

DeedsBCV\cite{deedsBCV1,deedsBCV2} is one of the top-performing traditional medical image registration software package. It uses rigid registration first, followed by a nonrigid registration.
These traditional methods usually perform very well in accuracy,
but their computing speed is very slow and cannot be used in real-time systems.

\subsection{Deep learning Registration Methods}
Many learning-based natural image registration methods\cite{gan1, gan2,gan3} are applied in automatic driving, augmented reality, and other fields.
They are often used to fuse images of different modalities, such as RGB and LiDAR images. For medical image registration, deep learning technology has been widely used in nonrigid registration tasks.
Reinforcement learning methods \cite{RL1, RL2} are used to simulate the traditional iterative registration method,
which is intuitive but costs too much time.
Supervised learning methods\cite{superviesd1, superviesd2} are used to predict a dense registration field,
but they require ground truth of the registration field, which is expensive and hard to obtain,
especially for nonrigid registration tasks.
So statistical appearance model (SAMs) \cite{superviesd_fake_data} is proposed to synthesize a registration field to generate simulation data and improve the accuracy of ground truth.
But it is difficult to set parameters and not robust in real scenarios.

Unsupervised methods do not need to label the images. One of the first papers using unsupervised methods for medical images registration tasks is \cite{basic}.
It uses a ConvNet\cite{convnet} to predict a registration field
and a grid sampler module proposed from STN\cite{stn} to warp the moving image to moved image.
The method is not fit for large deformation images registration because it calculates the registration field of image sub-regions
and does not consider cross-region registration.VoxelMorph replaces ConvNet with UNet and alleviates the above problem since it does not divide the image into sub-regions.
It optimizes the network through a similarity and a regularization loss function and achieves good performance on the brain MRI datasets.
But from the experimental results,
they perform poorly in dealing with larger-scale deformation chest images. The recursive-cascaded network\cite{rcn} (RCN) concatenates multiple voxelmorph or VTN to achieve a state-of-art performance. However, the inference time and memory increase significantly when more modules are cascaded. We introduce a multi-resolution refine model for large-scale deformation problems and the effectiveness of the method is verified by sufficient experiments.

\section{Methods}
We propose an unsupervised coarse to fine registration model,
which can optimize the registration field in different resolutions.
We find that even if the image pair has a large deformation at high resolution,
the difference will be much smaller if we reduce their resolution.
So we reduce the resolution to ignore small details.
The registration at low resolution is more like a global registration.
Then we gradually refine the registration field from coarse to fine in the registration model.

We use the same notations as \cite{VoxelMorph4}.
Let $M$ and $F$ denote the moving and fixed image separately,
$g$ is our registration model, $g_\theta(F, M)=\phi$ is the predicted registration field,
and $M(\phi)$ is moved image.
We use the 3D medical images in our research.
Both fixed images and moving images are three dimensional, single-channel, gray-scale images.

First, we concatenate $F$ and $M$ in one tensor and send them to the feature extraction path to generate multiple-scale features.
We predict the lowest resolution registration field $\phi_{n}$, a rotation matrix $R$, and a translation matrix $t$ using the highest abstract features.
Then we use the $R$ and $t$ to transform $\phi_{n}$ to imitate a wildly used rigid registration preprocess.
We refine the field from bottom to up in different resolution scales.
Then we use a grid sampler module proposed by STN\cite{stn} to sample the $M(\phi)$ from $M$.
Finally the  similarity loss will be calculated comparing the $M(\phi)$ with $F$ and a regularization loss will be calculated by $\phi_{0}...\phi_{n}$.
The loss will be back-propagated through the network to adjust parameters using stochastic gradient descent algorithm.

\subsection{Registration Model}
As shown in Figure \ref{fig:network}, the registration network has three paths,
including a feature extraction down-sampling path,
a pooling down-sampling path, and a refine up-sampling path.
And the feature and pooled image are attached to the fine-tuning module in the same registration stage using skip connection.

Feature extraction path is used to extract features in different resolutions.
We use standard convolution followed by Leaky Relu activation function and Batch Normal layer as basis module. It also can be replaced with any effective feature extraction network such as ResNet\cite{resnet}, DenseNet\cite{densenet}, etc.
The input in our experiment is two channels, which are combined $F$ and $M$ in one tensor. The feature in the coarsest layer has 64 channels by default. We first generate transform flow $\phi_{n}$ by two 3D convolution layers which reduce the channel from 64 to 3. Then transform the $\phi_{n}$ by rigid block described in section~\ref{sec: rigidblock}.

A pooling path is introduced to evaluate the registration results of each stage and provide auxiliary information. The reason why we choose average-pooling rather than max-pooling is that max-pooling will magnify the local maximum point,
which will cover too much area in lower resolution image.

Refine path is used to correct the registration field at different resolution.
First, we generate a primary registration field at the lowest resolution.
We optimize the higher resolution layer when the lower resolution registration is acceptable. The model refines the result step by step from bottom to up.

In addition, most registration algorithms contain rigid registration as preprocessing.
We design a rigid block to predict rotation matrix $R$ and translation matrix $t$.
We embed it into the model and optimize it by unsupervised method,
which can reduce the memory cost and improve the registration effect.

\begin{figure}[!tbp]
\includegraphics[width=0.9\linewidth]{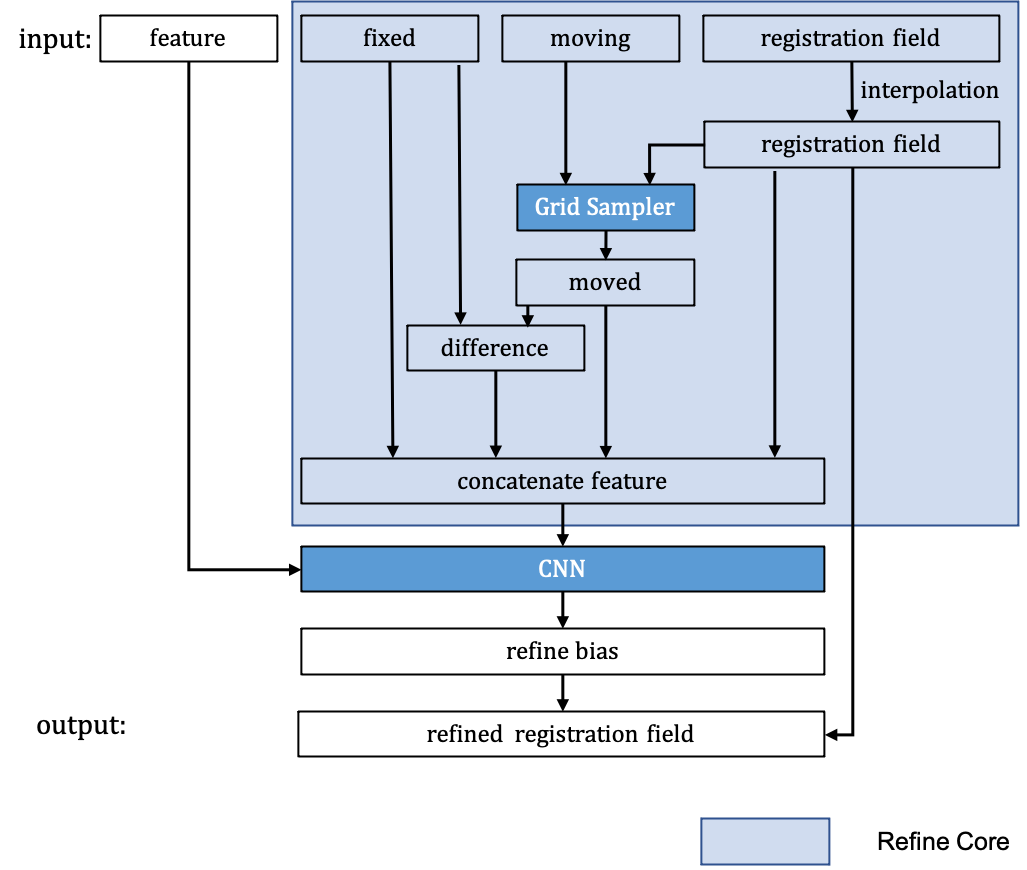}
\caption{Refine block.
The input includes four components, feature from feature extraction path,
fixed image and moving image from pooling path,
and registration field in smaller resolution from previous stage.
The area enclosed by the blue box is $Refine \ Core$, also called as $RC$.}
\label{fig:refine block}
\end{figure}

\subsection{Refine Block}
Refine block is the basic module of the refine path, as shown in Algorithm \ref{algorithm:refine}.
The input of the refine block consist of four parts,
moving image $M_i$, fixed image $F_i$, $Feature_i$ in the same stage,
and registration field $\phi_{i-1}$ of previous stage.
$M_i$ and $F_i$ are pooled through the average pooling path to get the same resolution with current refine stage.
$Feature_i$ is connected to the refine block using a skip connection.

\begin{figure}[h]
\includegraphics[width=1.0\linewidth]{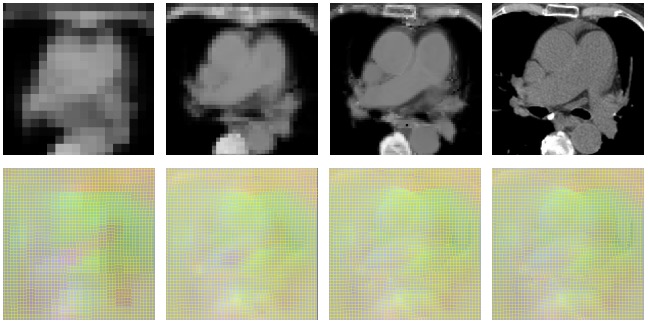}
\caption{Images of different resolutions from coarse(left) to fine(right) and registration fields generated by each refine block.}
\label{fig:coarse2fine}
\end{figure}

First, we interpolate the $\phi_{i-1}$ to $\Tilde{\phi_{i-1}}$ which has the same resolution with $F_i$ using triple linear interpolation.
Then we warped $M_i$ with $\Tilde{\phi_{i-1}}$ to get a warped moving image $M(\phi)_i$,
this result indicates the registration performance of previous stage, then we should refine it in the current stage.
We make $M(\phi)_i$ minus $F_i$ and get a difference tensor $D_i$,
then we use it together with $M(\phi)_i$, $F_i$, $Feature_i$ and $\Tilde{\phi_{i-1}}$ as a grouping feature to predict a refined registration field $\phi_i$ using standard convolution layer followed by a Leaky Relu activation function and Batch Normal layer.

The main component of refine block is the refine core, as shown in Figure \ref{fig:refine block}.
If we remove this part, it will act as a standard UNet\cite{UNet} up-sampling path
and we have demonstrated the necessity of refine core using an ablation experiment in Section \ref{tab:ablation}.

\begin{algorithm}
    \caption{$Refine\ Block$}
    \begin{algorithmic}[1]
        \Require $M_i, F_i, Feature_i, \phi_{i-1}$ 
        \Ensure $\phi_i$
        \State $g_{sample} \gets Grid\ Sampler$\Comment{$pytorch \ function$}
        \State $g_{conv} \gets Conv\ layer$
        \State $\Tilde{\phi_{i-1}} \gets Interpolate(\phi_{i-1})$
        \State $M(\phi)_i \gets g_{sample}(M_i, \Tilde{\phi_{i-1}})$
        \State $D_i \gets M(\phi)_i-F_i$
        \State $\Tilde{Feature_i} \gets cat(Feature_i,D_i,F_i,M(\phi)_i,\Tilde{\phi_{i-1}})$
        \State $\phi_i \gets \Tilde{\phi_{i-1}}+g_{conv}(\Tilde{Feature_i}) $
        \State \Return{$\phi_i$}
    \end{algorithmic}
    \label{algorithm:refine}
\end{algorithm}

In the final stage, we combined all stage features to achieve a multiple scale perception effects by a CNN module. The CNN module is composed of four 3D convolution layers with batch normalization and leaky relu activation, the output channel for each layer are 64, 32, 8 and 3 respectively.

\subsection{Rigid Block}
\label{sec: rigidblock}
A general idea of the nonrigid registration algorithm is to do rigid registration first, and then do nonrigid registration based on the previous rigid transformation matrix.
We took the same approach, but the difference is that we chose to integrate them into one network for the sake of convenience and simplicity.

First, we pass the output of the feature extraction path through two 3D convolution layers, which output channel is 64 and 16 respectively. Then we use a max function on all channels to gather the maximum value to reduce the computation and improve the abstractness. We reshape the feature to a 1-D vector of length $(H*W*L)/(16*16*16)$ and predict the rotation $R\in\mathbb{R}^{3\times3}$ and translation $t\in\mathbb{R}^{1\times3}$ using two independent branches. Each branch consists of three fully connected layers. Finally, we transform coarse registration field $\phi_{n}$ using $R$ and $t$.

We choose to transform the registration field instead of the image, to prevent the feature and the image point from not corresponding in the refine path. We performed ablation experiments to verify the necessity of this block. The results are shown in Table \ref{tab:ablation}.

\subsection{Loss Function}
We use the same sample method as STN,
which used a grid to sample the moving image using the predicted registration field to achieve registration.
Each point of the output tensor has three dimensions,
which represent the coordinates X, Y, Z of the sampling points on the moving image.
To meet the needs of different input sizes,
We need to normalize the output coordinates and map them to [-1,1],
and this requires the output of the network to be small enough to prevent cross-border,
so the loss function includes a regularization component $L_{range}$ (\ref{loss_range}) to restrict the range of the output. Considering the actual situation, the sampling coordinates of adjacent points should be close.
So we add $L_{smooth}$ (\ref{loss_smooth}) to our loss function to restrict the smoothness of the registration field.

We use Mean Square Error(MSE) to measure the similarity of $M$ and $F$ (\ref{loss_similarity}).
Many algorithms use cross-correlation(CC) as a function of similarity measurement.
However, in chest organ registration tasks,
rib, and other high gray value structures are often the background,
and the calculation of cross-correlation is based on the product of corresponding points,
which will lead to the loss function excessively concentrated in the background,
resulting in poor effect and even unable to converge.

Above all, our loss function is:
\begin{align}
    \label{loss}
    &L(F,M(\phi),\phi)=\sum_{s=0}^S(L_{sim}(F,M(\phi))+\lambda L_{reg}(\phi)) \\
    \label{loss_similarity}
    &L_{sim}(F, M(\phi))= \frac{1}{N}\sum_{p \in \Omega}^N(M_p-F_p)^2\\
    \label{loss_reg}
    &L_{reg}(\phi)= \alpha L_{range}+\beta L_{smooth} \\
    \label{loss_range}
    &L_{range} = \frac{1}{N}\sum_{p \in \Omega}^N|\phi_p| \\
    \label{loss_smooth}
    &L_{smooth} = \frac{1}{N}\sum_{p \in \Omega}^N|\nabla \phi(p)|
\end{align}
    
where S is the total stage number of the network, $\lambda$ is regularization parameter,
$\alpha$ and $\beta$ are weights of different component of regularization.

\section{Experiment}

\subsection{Dataset}
Our algorithm is designed for chest CT registration, and there is no public dataset available for this registration task. We choose SegTHOR\cite{SegTHOR}, a widely used public dataset for chest multiple organ segmentation as our test dataset. So that performance can be compared with other registration methods. SegTHOR consists of 60 chest CT scans and more details are available in their paper.

We try to train and test our model on SegTHOR dataset but find the quantity is insufficient to train an unsupervised registration model. So we build a private dataset consists 682 chest CT scans from the hospital. The in-plane resolution varies between 0.35 mm and 0.98 mm per pixel and the z-resolution fluctuates between 0.6 mm and 2.0mm. We randomly selected 48 chest CT scans from the private dataset as test dataset and the rest is training dataset. We use the public dataset SegTHOR to test the results. We cut out the volume which contains the heart, aorta, trachea and esophagus from chest CT and truncates the gray level to the mediastinum window, and normalize to $[-1, 1]$. The preprocessed images are resized to 128x128x128 for training and testing.

We also use the LPBA40 dataset for comparison between our method with the large deformation registration method LapIRN\cite{lapirn}. We split the LPBA40 dataset into 28, 2 and 10 volumes for training, validation and test sets. The images are center cropped to 144x192x160. No data augmentation is implemented.

\subsection{Details}
Experiments are implemented on Intel Xeon(E5-2685) CPU and NVIDIA TITAN RTX GPU.
We use Adam\cite{adam} as optimizer, and the learning rate is $10^{-4}$.
The average dice score of all chest organs is used to measure the registration performance of our model.
We train our model stage-wise from coarse to fine. We use MSE as our similarity loss function, so we don\rq{t} use standardization.

In our experiment, $\lambda$ in Equation\ref{loss} is $10^3$,
$\alpha$ and $\beta$ in Equation\ref{loss_reg} is $10$ and $10^2$.
These parameters should be adjusted according to the size of the input tensor and the ratio of the foreground to the background.

\subsection{Baselines}
We compare our methods with two state-of-the-art traditional software package ANTs\cite{Ants} and deedsBCV\cite{deedsBCV1,deedsBCV2}.
We also made a comparison with the deep learning method VoxelMorph\cite{VoxelMorph4} and RCN\cite{rcn}. We use the VoxelMorph-2 structure described in the paper. The model channel is doubled, i.e., the maximum channel for one layer is 64.  For RCN\cite{rcn}, a 3-cascade VTN is used, which has a comparable inference time and memory. Additionally, we compared our methods with LapIRN on private and LPBA40 datasets. LapIRN is designed for large deformation registration. We use the code provided by the author and change the start channel to 7. We do not use the displacement fields version since it cannot guarantee to be smooth and locally invertible. All the comparisons used the same data loader and dice calculation. 
  
\subsection{Results}
We evaluate our methods on the public dataset SegTHOR (Table \ref{tab:dice_on_segthor}) and private dataset (Table \ref{tab:dice_on_private}).
Our method is better than the deep learning method VoxelMorph and traditional algorithm ANTs.
It is a little better than one of the top-performing traditional algorithm deedsBCV. We also compare our model with LapIRN on the private and LPBA40 datasets(Table \ref{tab:dice_lapirn}). Our model was able to achieve slightly better results than the LapIRN, validating its effectiveness. In addition, we achieve an exponential improvement in computing speed compared with traditional
registration methods and our model can be used in real-time systems (Table \ref{tab:runtime}).

The z-axis resolution of SegTHOR dataset is between 2.00 mm and 3.70 mm per pixel,
which is different from that of our private testing dataset and fixed image (between 0.60 mm and 2.00 mm per pixel).
So the test results are different between the private test dataset and SegTHOR dataset. But in both tests we achieve the-state-of-art results.

\begin{figure}[!tbp]
\includegraphics[width=0.9\linewidth]{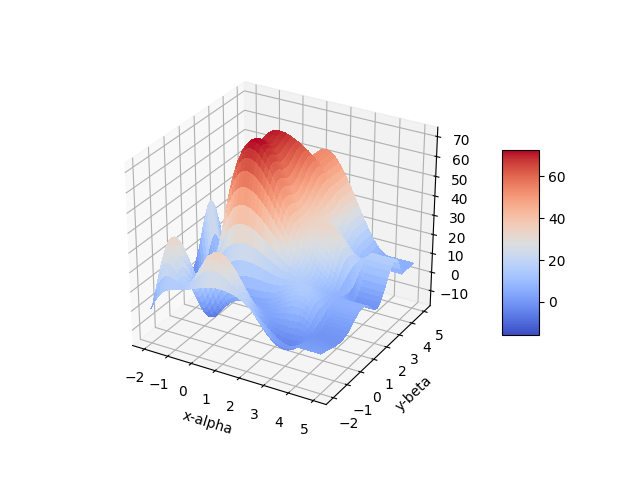}
\caption{Experiments of different combination of $\alpha$(x-axis) and $\beta$(y-axis).
x-axis is $\log \alpha$.
y-axis is $\log \beta$.
z-axis is dice scores.}
\label{fig:hyper-parameter}
\end{figure}

\begin{table}[h]
    \centering
    \caption{Dice scores on SegTHOR.}
    \label{tab:dice_on_segthor}
    \begin{tabular}{lccccc}
        \hline
        \ & Heart& Aorta& Trachea& Esophagus& Mean\\
        \hline
        $VoxelMorph\cite{VoxelMorph4}$& 67.52& 35.12& 16.82& 8.13& 31.90\\
        $Ants\cite{Ants}$& 74.88& 57.47& 54.32& 38.81& 56.37\\
        $deedsBCV\cite{deedsBCV1,deedsBCV2}$& 80.32& 75.15& 66.71& 51.19& 68.34\\
        $RCN\cite{rcn}$& 84.61& 71.59& 70.26& 45.58& 68.21\\
        $Ours$& 84.50& 72.98& 71.03& 45.16& 68.42\\
        \hline
    \end{tabular}
    \bigskip

    \caption{Dice scores on Private Dataset.}
    \label{tab:dice_on_private}
    \begin{tabular}{lccccc}
        \hline
        \ & Heart& Aorta& Trachea& Esophagus& Mean\\
        \hline
        $VoxelMorph\cite{VoxelMorph4}$& 81.36& 66.95& 13.75& 17.26& 44.83\\
        $Ants\cite{Ants}$& 74.41& 69.68& 76.61& 48.13& 67.21\\
        $deedsBCV\cite{deedsBCV1,deedsBCV2}$& 80.75& 83.55& 71.83& 53.32& 72.36\\
        $RCN\cite{rcn}$& 83.8& 79.24& 74.69& 50.63& 72.12\\
        $Ours$& 83.38& 79.48& 78.35& 50.32& 72.88\\
        \hline
    \end{tabular}
    \bigskip
    
    \caption{Dice comparison of LapIRN and LDRNet.}
    \label{tab:dice_lapirn}
    \begin{tabular}{lccccc}
        \hline
        \ & LDPA40& Private Dataset\\
        \hline
        $LapIRN\cite{lapirn}$& 70.89& 72.34\\
        $Ours$& 71.21& 72.88\\
        \hline
    \end{tabular}
    \bigskip
    
    \caption{Dice scores of Ablation Experiments on SegTHOR.}
    \label{tab:ablation}
    \begin{tabular}{lccccc}
        \hline
        \ & Heart& Aorta& Trachea& Esophagus& Mean\\
        \hline
        $Ours \ w/o \ RC$& 70.15& 47.19& 9.83& 11.24& 34.60\\
        $Ours \ w/o \ rigid$& 74.13& 60.67& 61.81& 38.54& 58.79\\
        $Ours \ w/o \ L_{range}$& 79.41& 43.49& 0.00& 4.69& 31.90\\
        $Ours \ w/o \ L_{smooth}$& 55.19& 15.17& 4.76& 5.59& 20.18\\
        $Ours$& 84.50& 72.98& 71.03& 45.16& 68.42\\
        \hline
    \end{tabular}
\end{table}

\begin{table*}[!tbp]
    \centering
    \caption{Visualization result on SegTHOR.
    Gray image registration result(row 1-2) shows the fitting ability of various methods.
    Mask image registration result(row 3-4) shows the regional continuity of various methods.
    The mask includes four parts: heart(green), aorta(yellow), trachea(blue) and esophagus(red).}
    \label{tab:visualize}    
    \begin{tabular}{ccccccc}

        \includegraphics[width=0.14\linewidth]{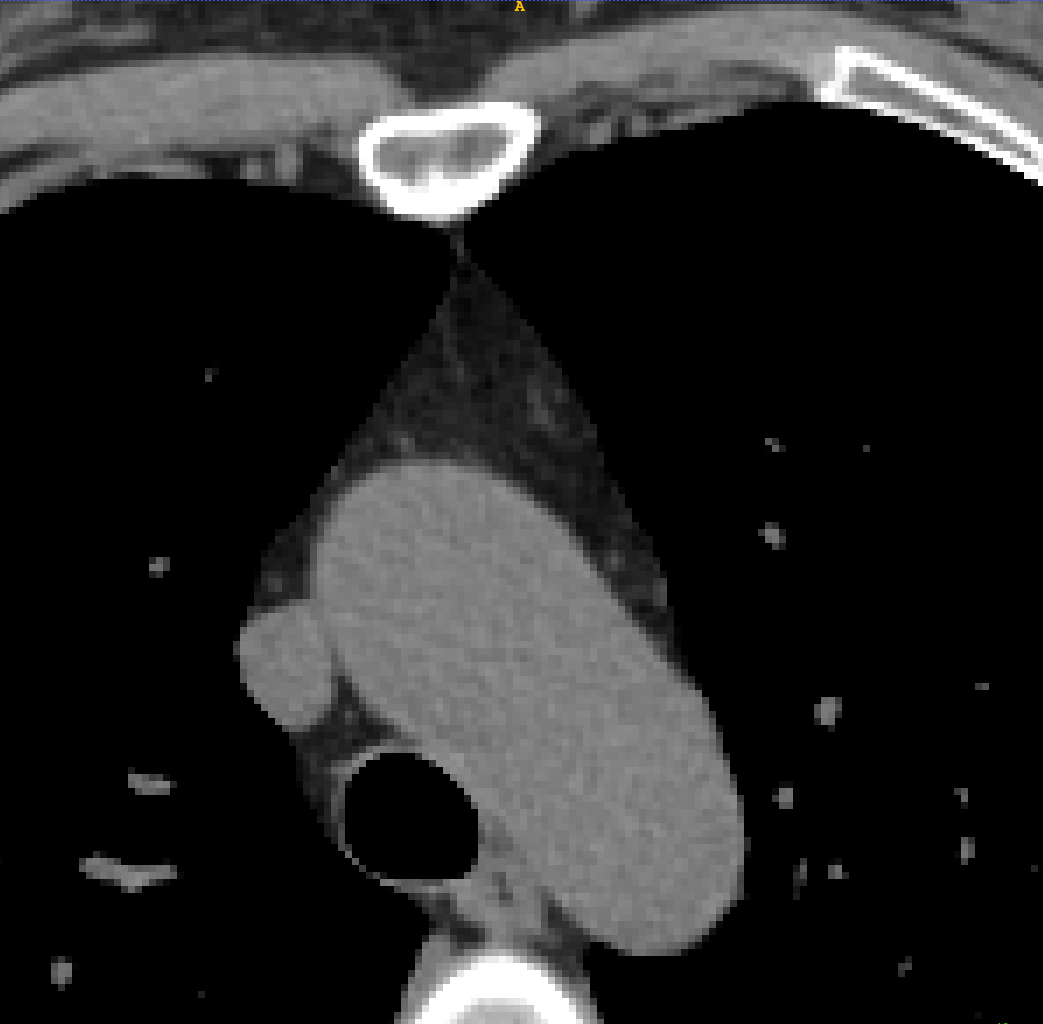}&
        \includegraphics[width=0.14\linewidth]{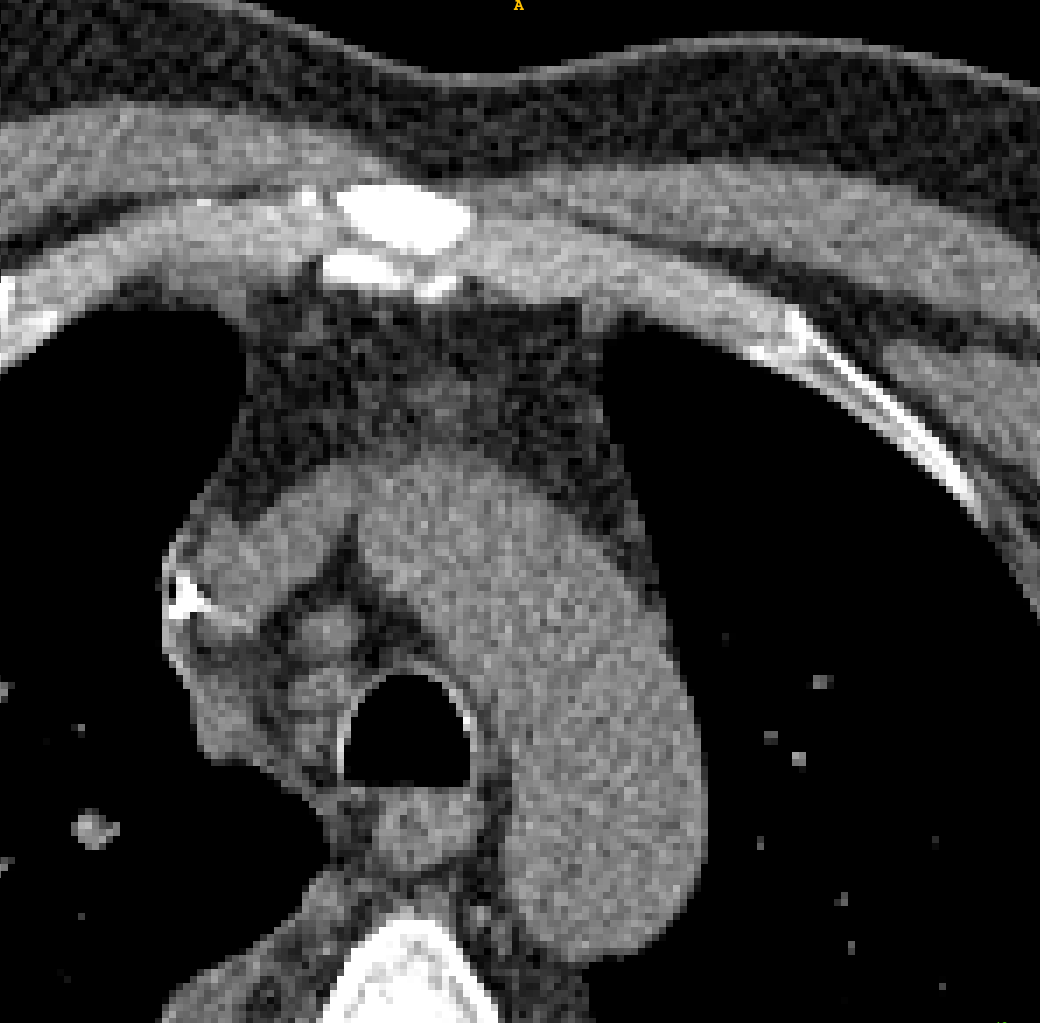}&
        \includegraphics[width=0.14\linewidth]{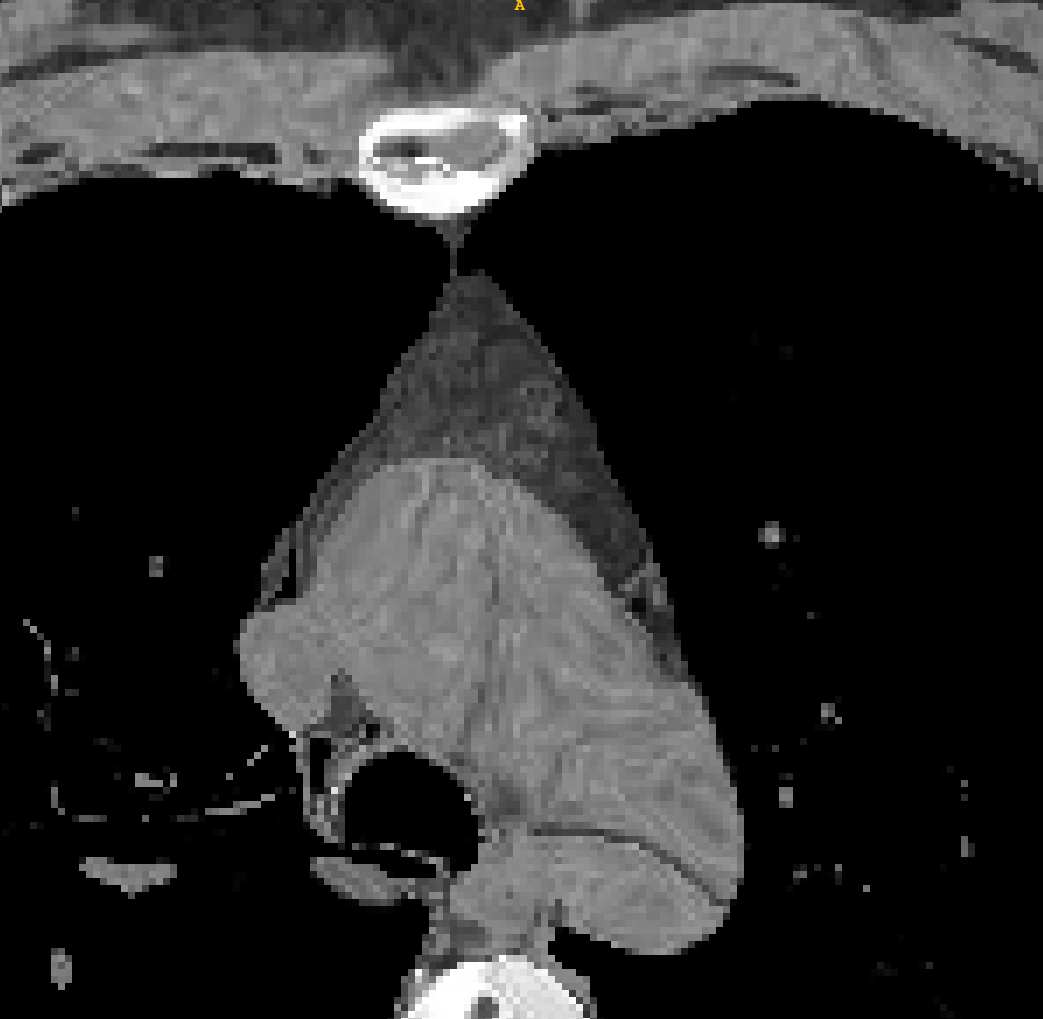}&
        \includegraphics[width=0.14\linewidth]{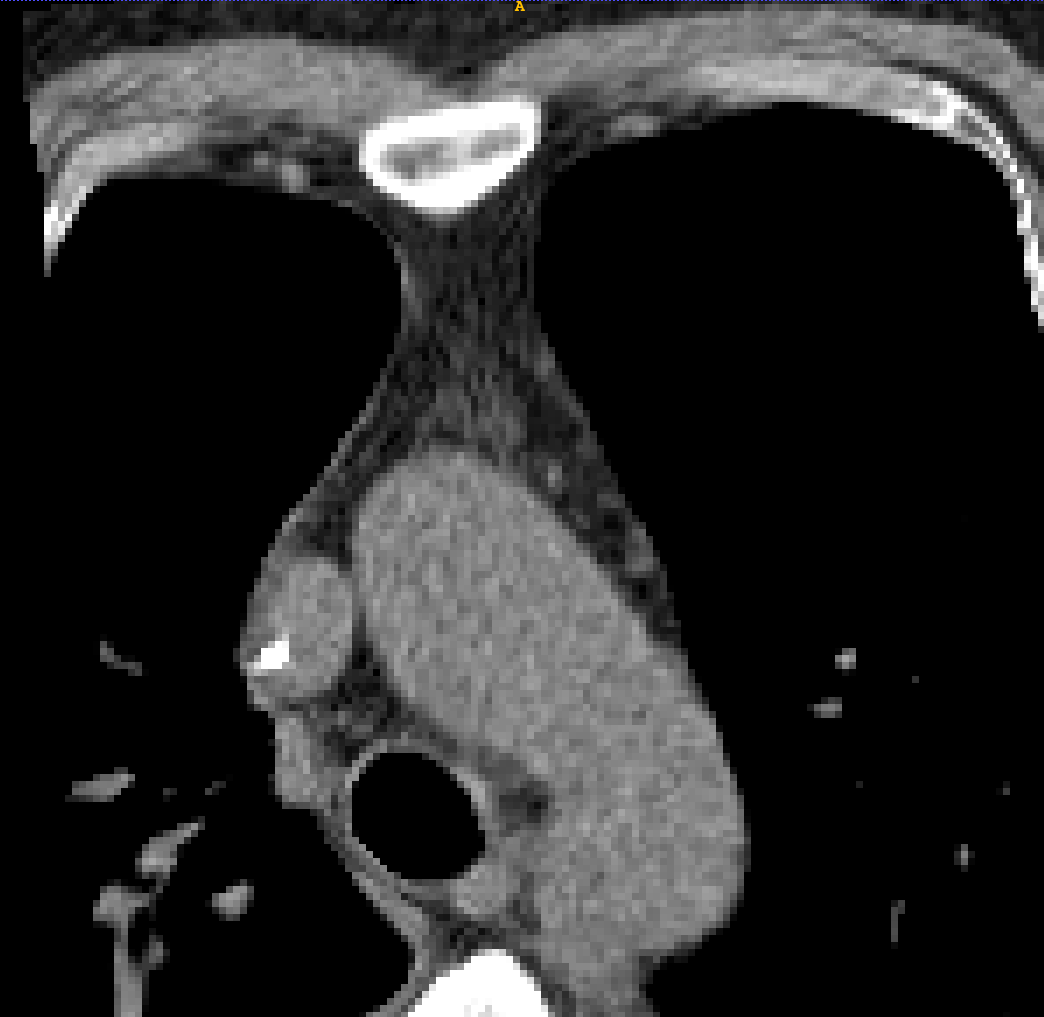}&
        \includegraphics[width=0.14\linewidth]{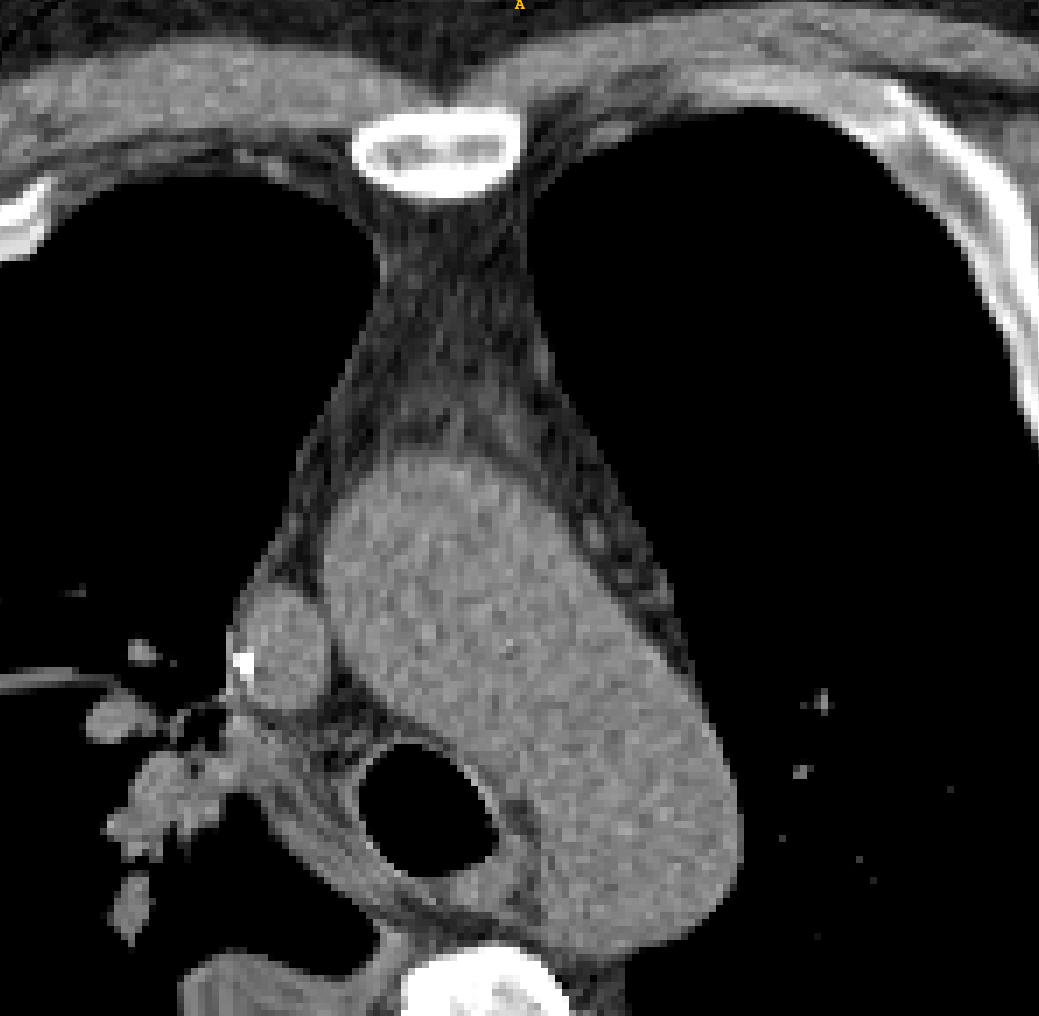}&
        \includegraphics[width=0.14\linewidth]{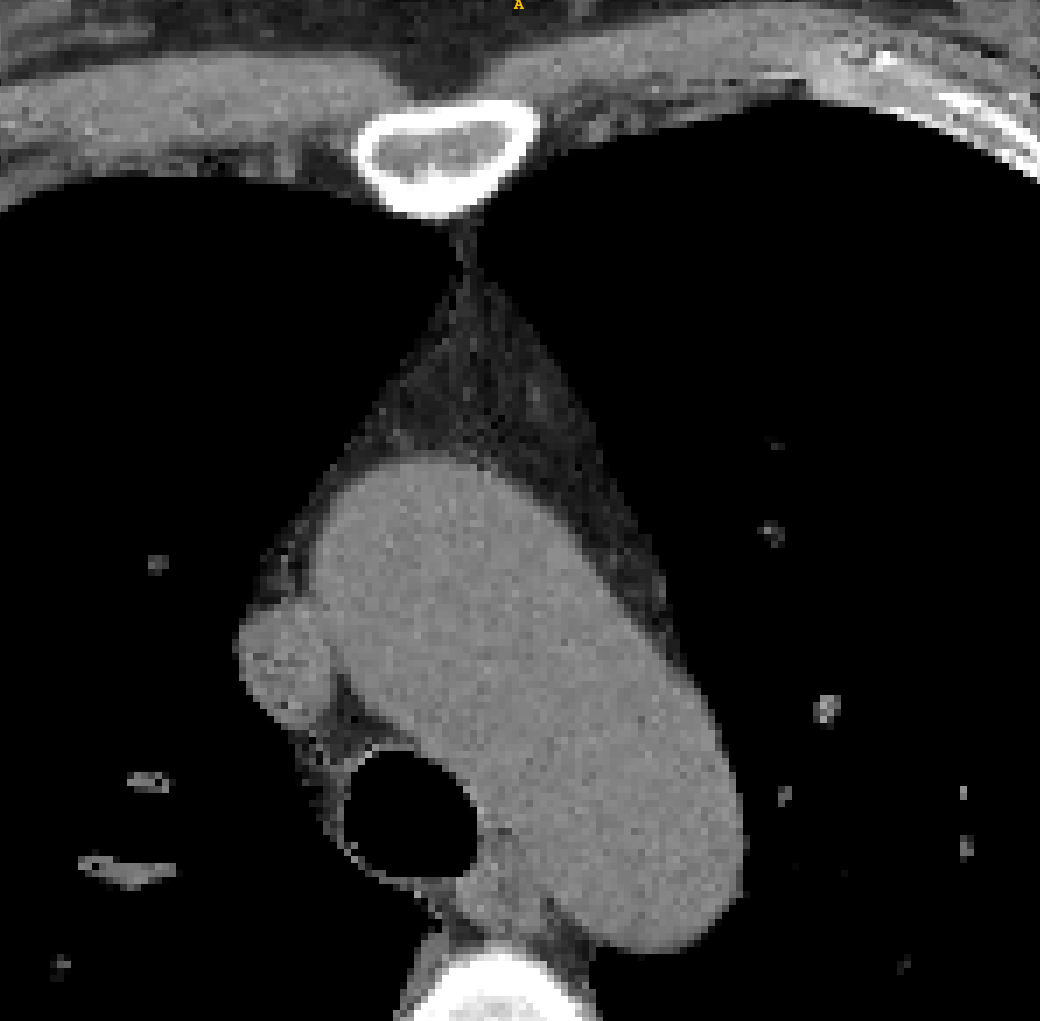}\\

        \includegraphics[width=0.14\linewidth]{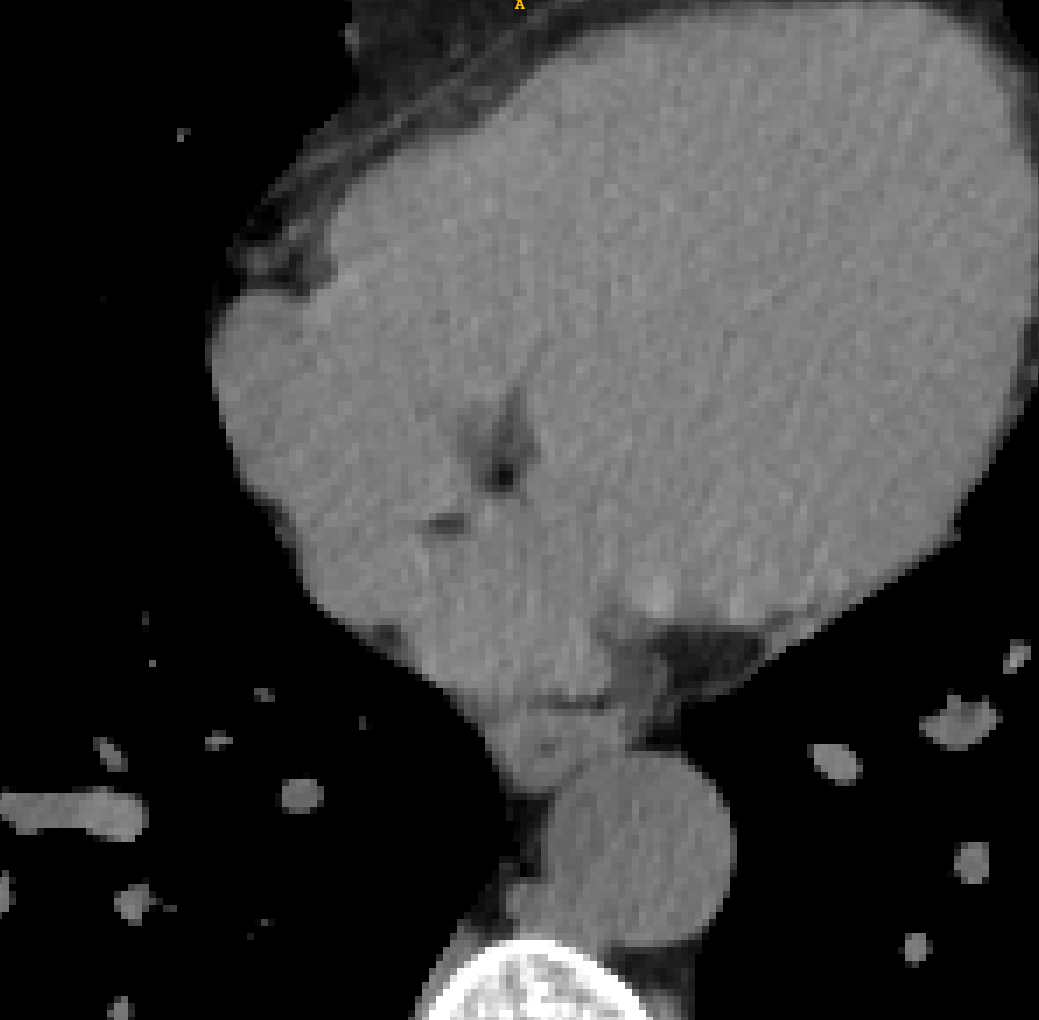}&
        \includegraphics[width=0.14\linewidth]{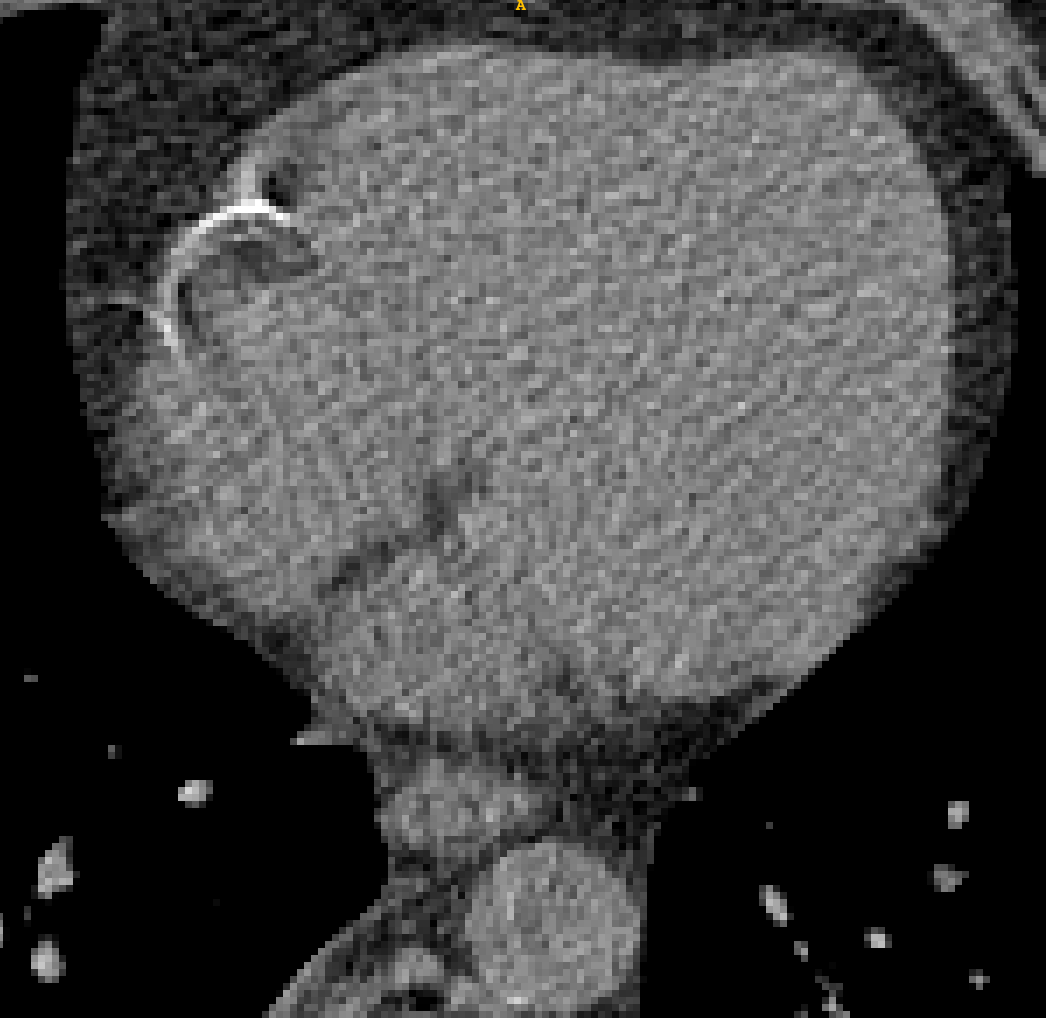}&
        \includegraphics[width=0.14\linewidth]{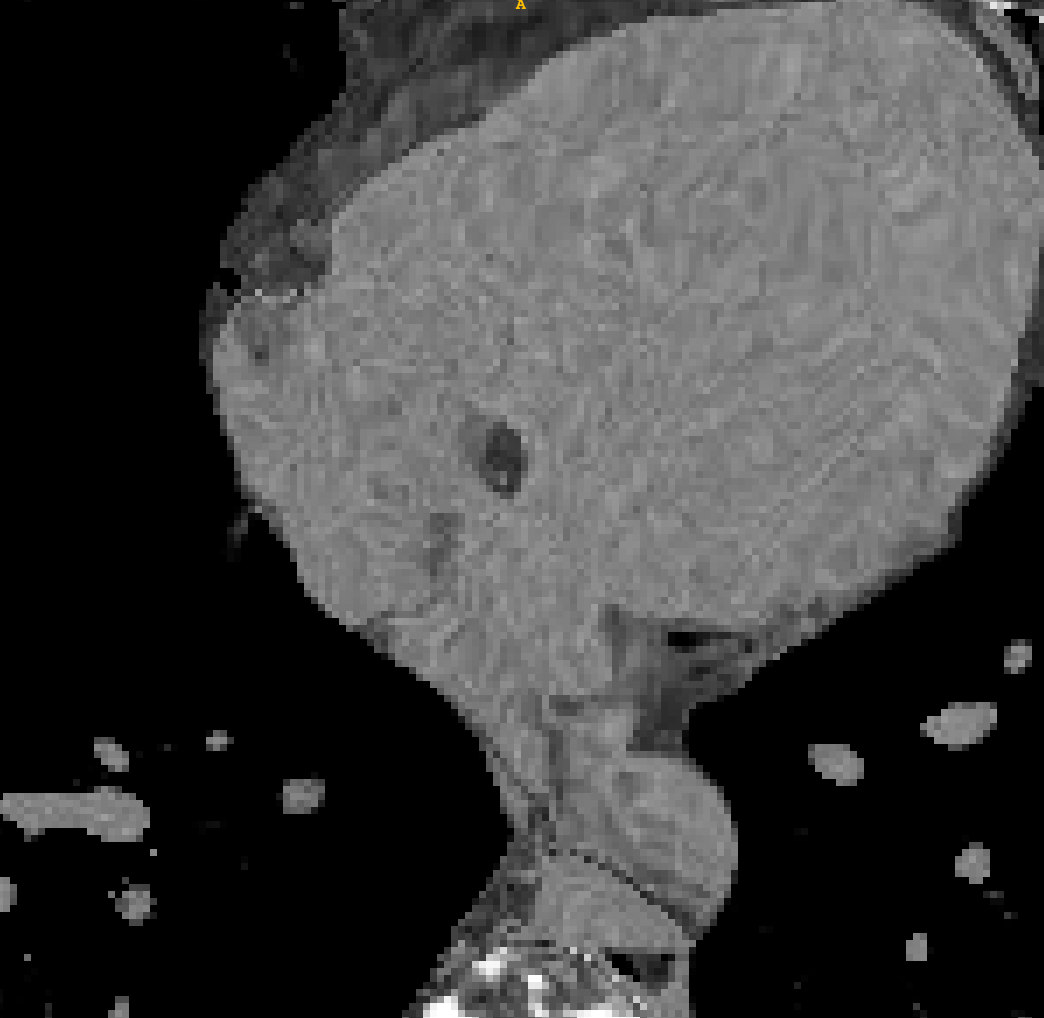}&
        \includegraphics[width=0.14\linewidth]{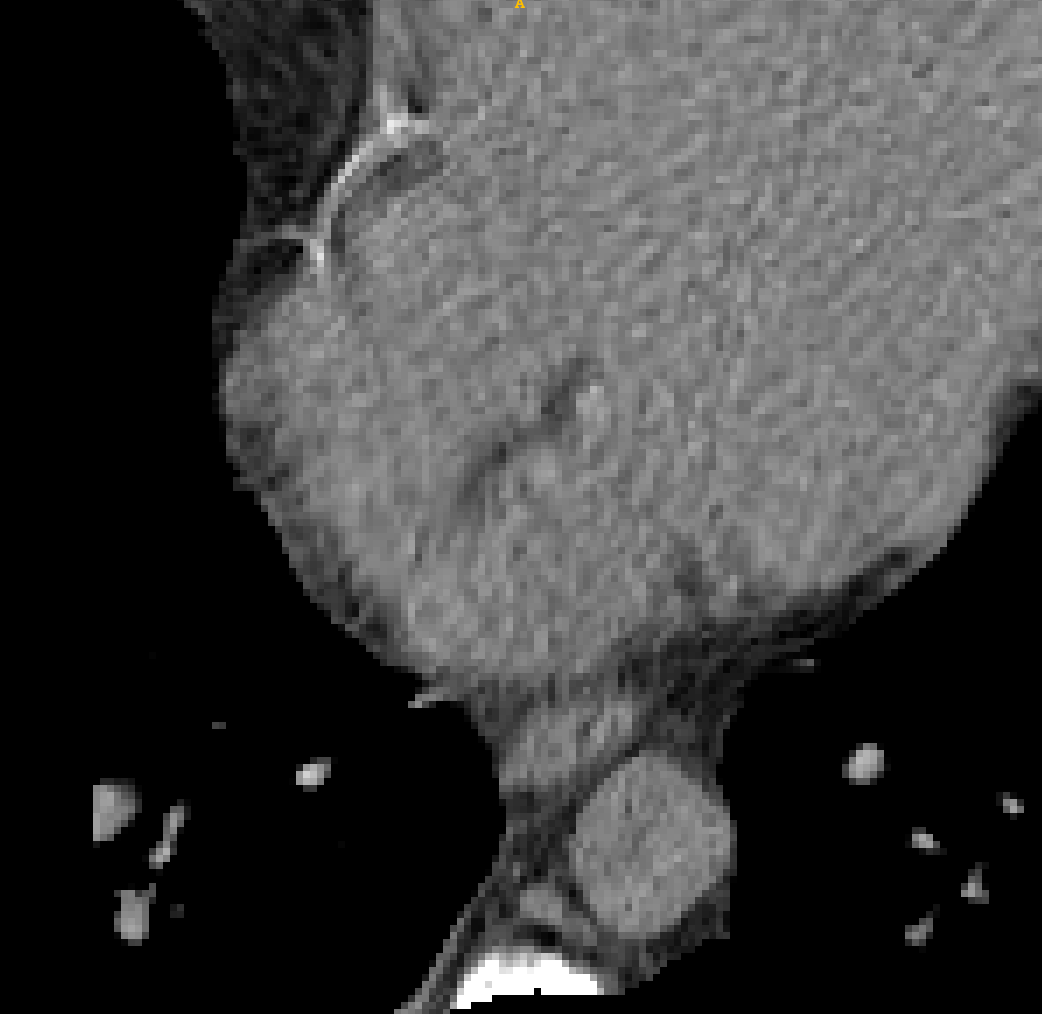}&
        \includegraphics[width=0.14\linewidth]{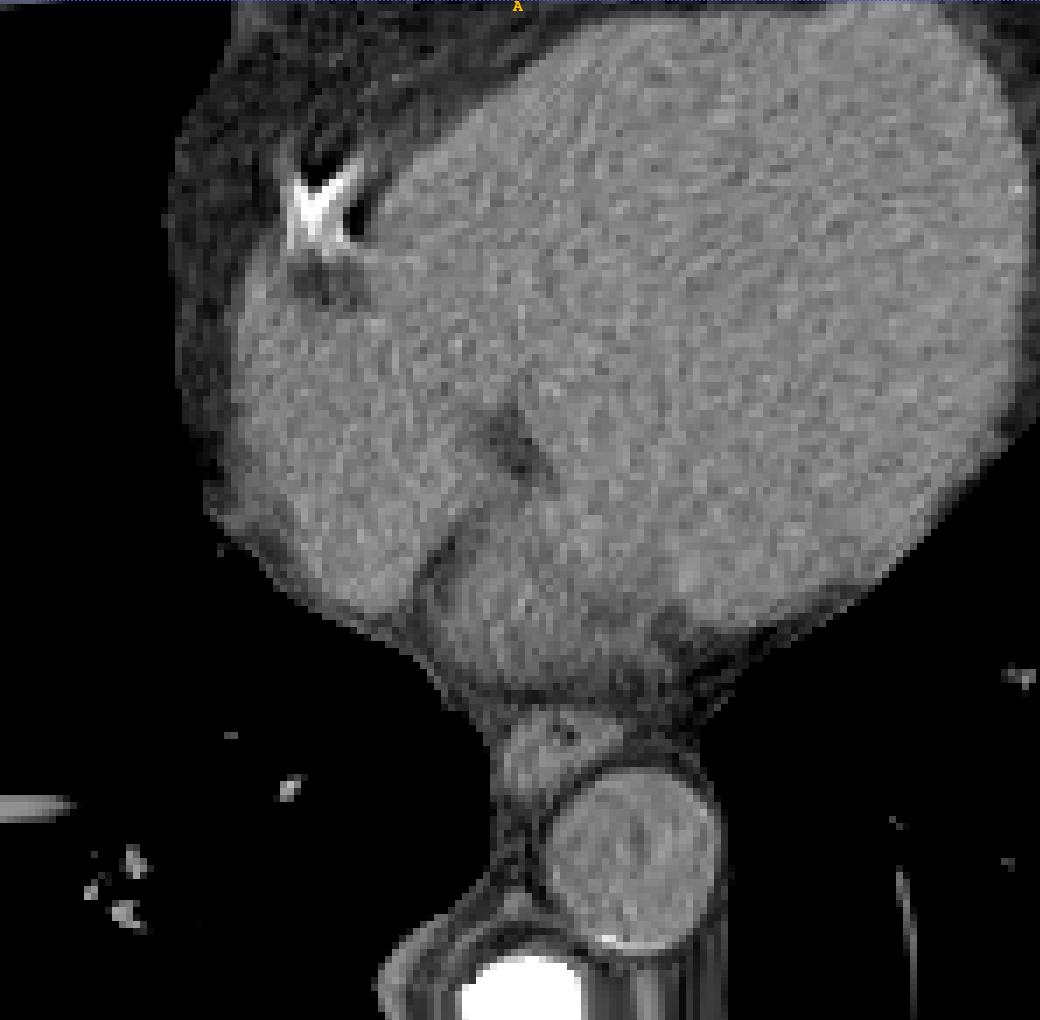}&
        \includegraphics[width=0.14\linewidth]{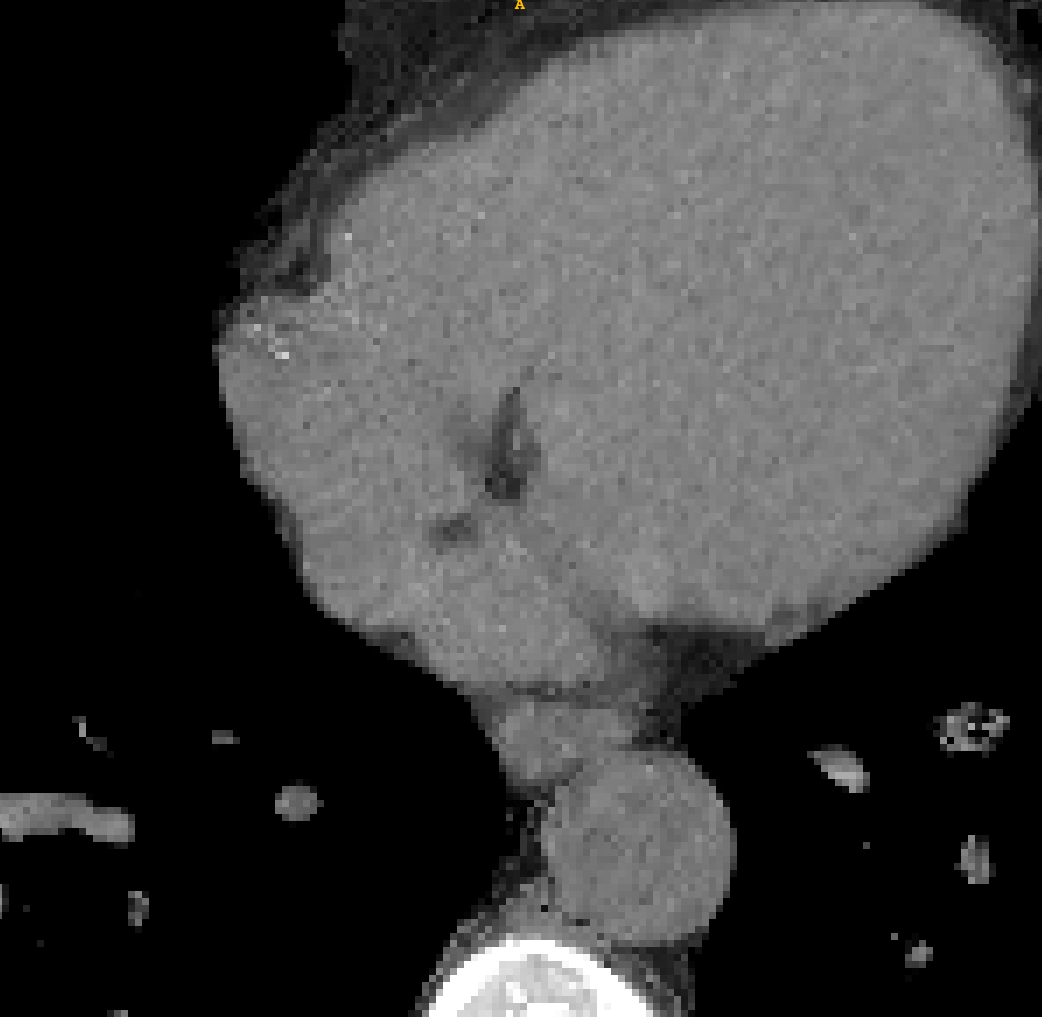}\\

        \includegraphics[width=0.14\linewidth]{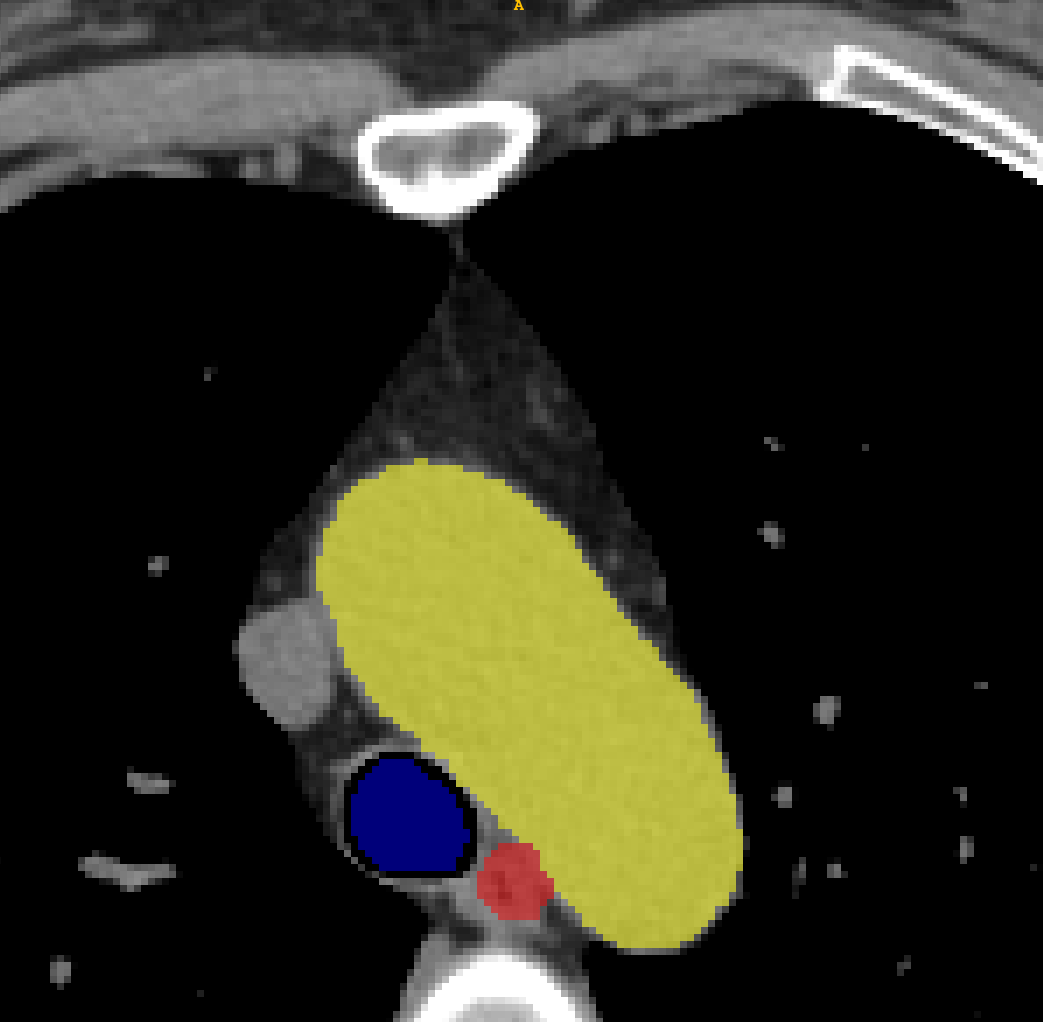}&
        \includegraphics[width=0.14\linewidth]{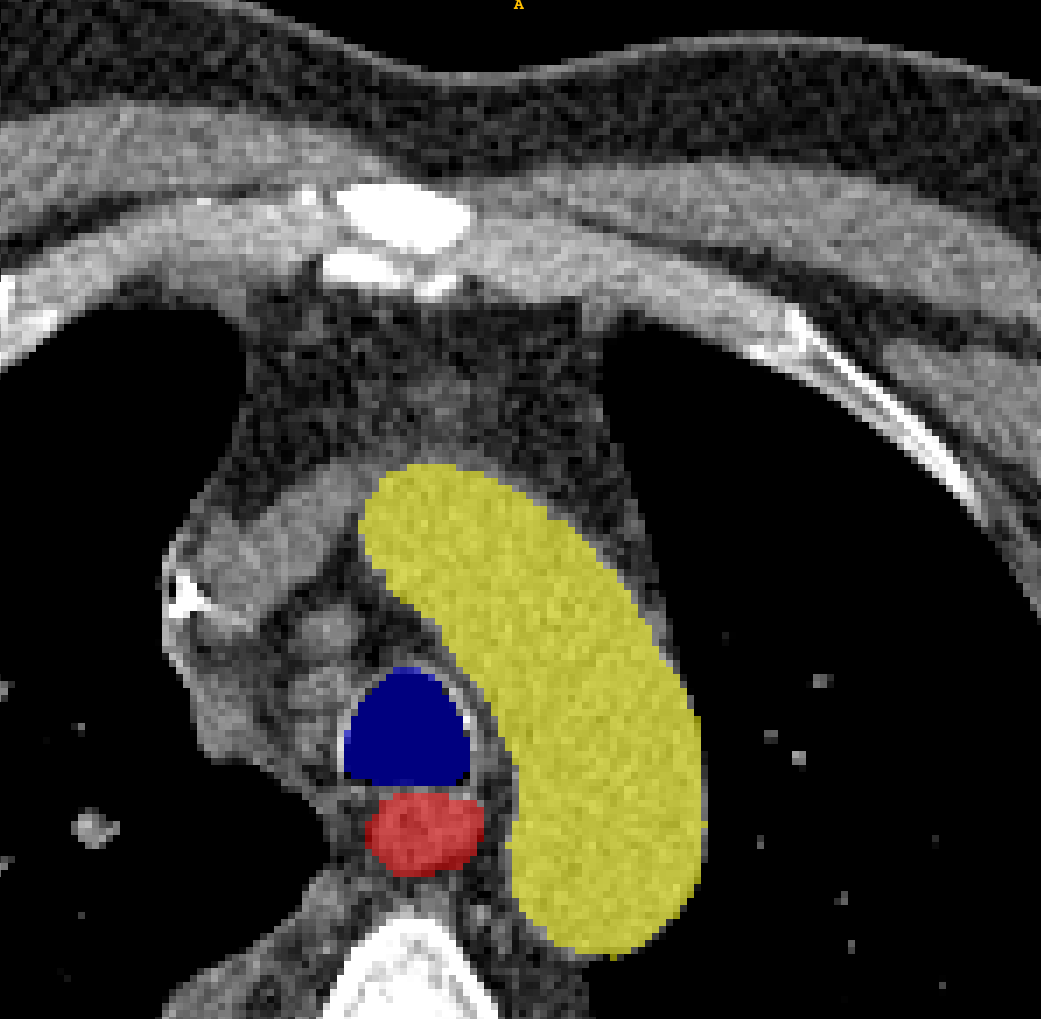}&
        \includegraphics[width=0.14\linewidth]{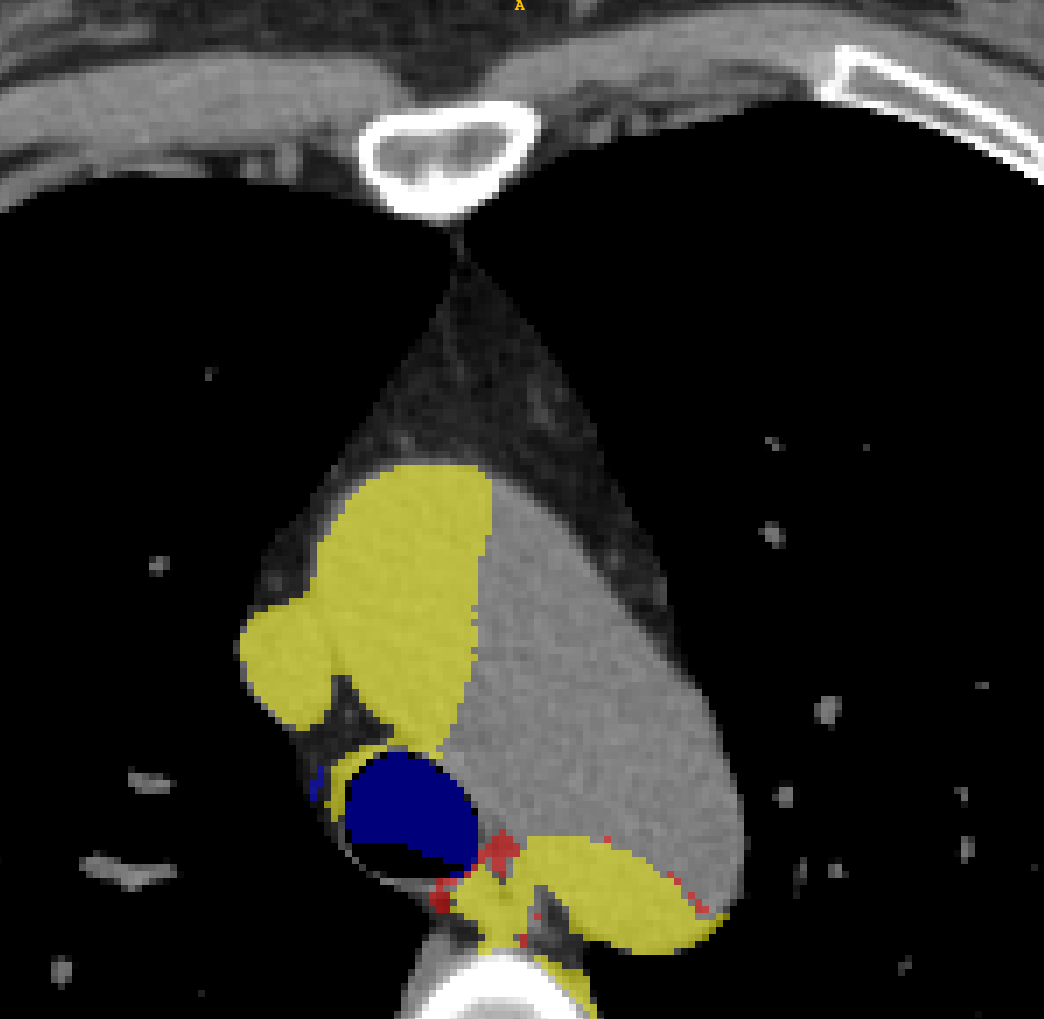}&
        \includegraphics[width=0.14\linewidth]{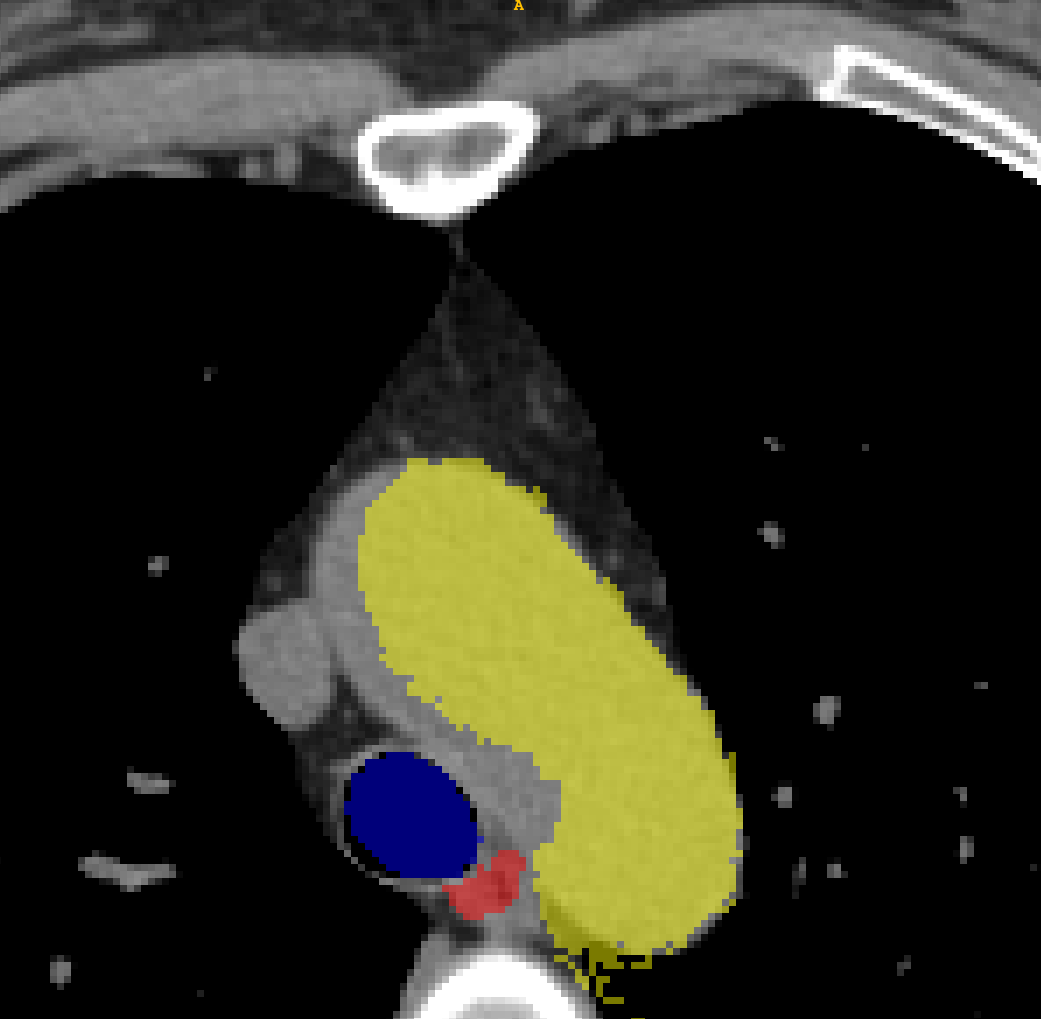}&
        \includegraphics[width=0.14\linewidth]{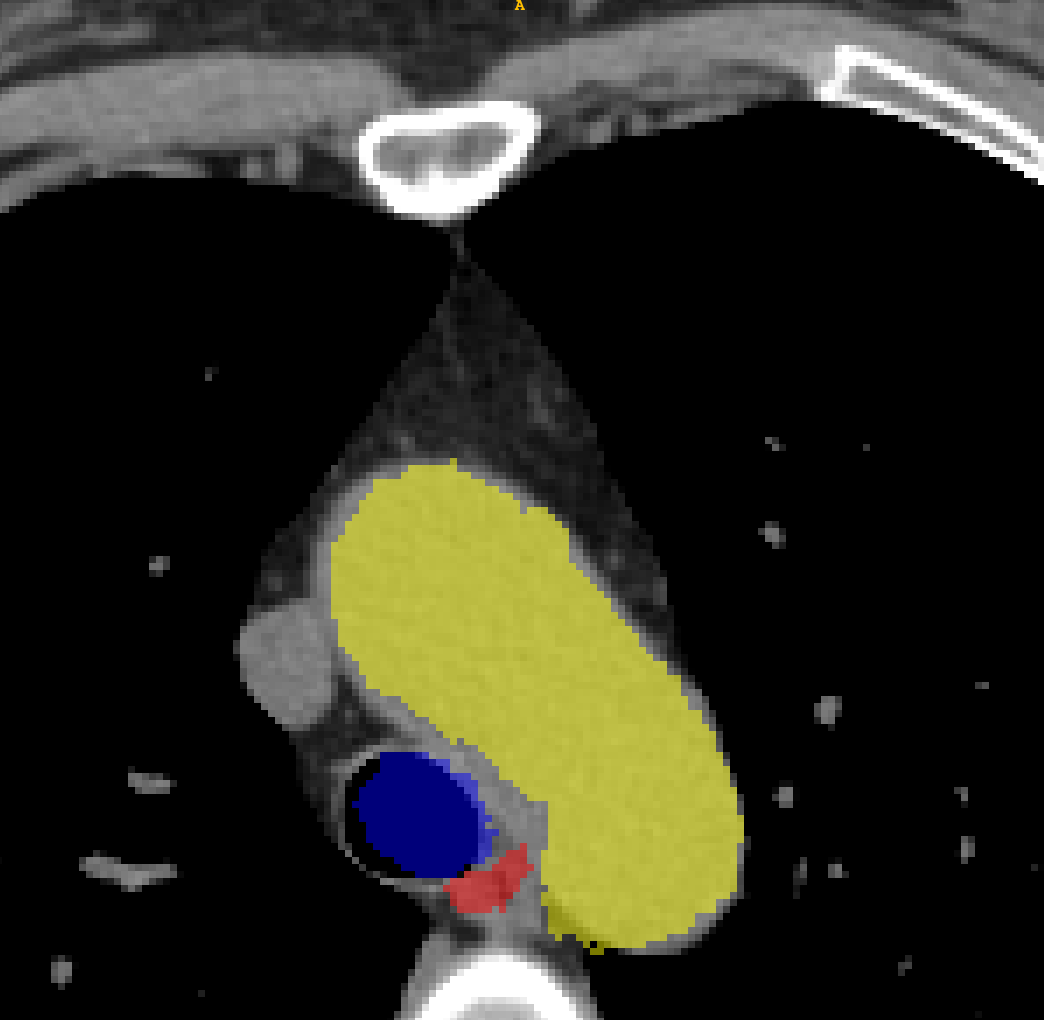}&
        \includegraphics[width=0.14\linewidth]{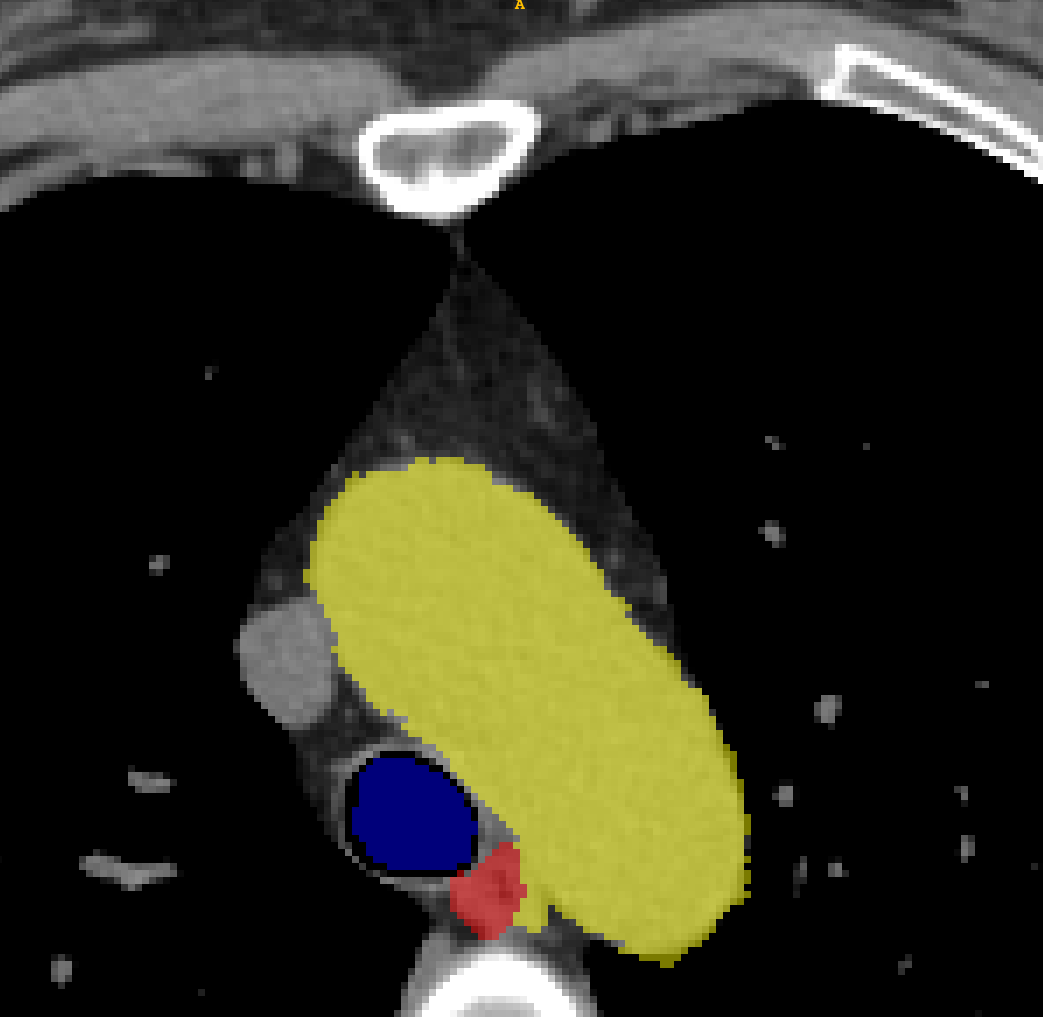}\\

        \includegraphics[width=0.14\linewidth]{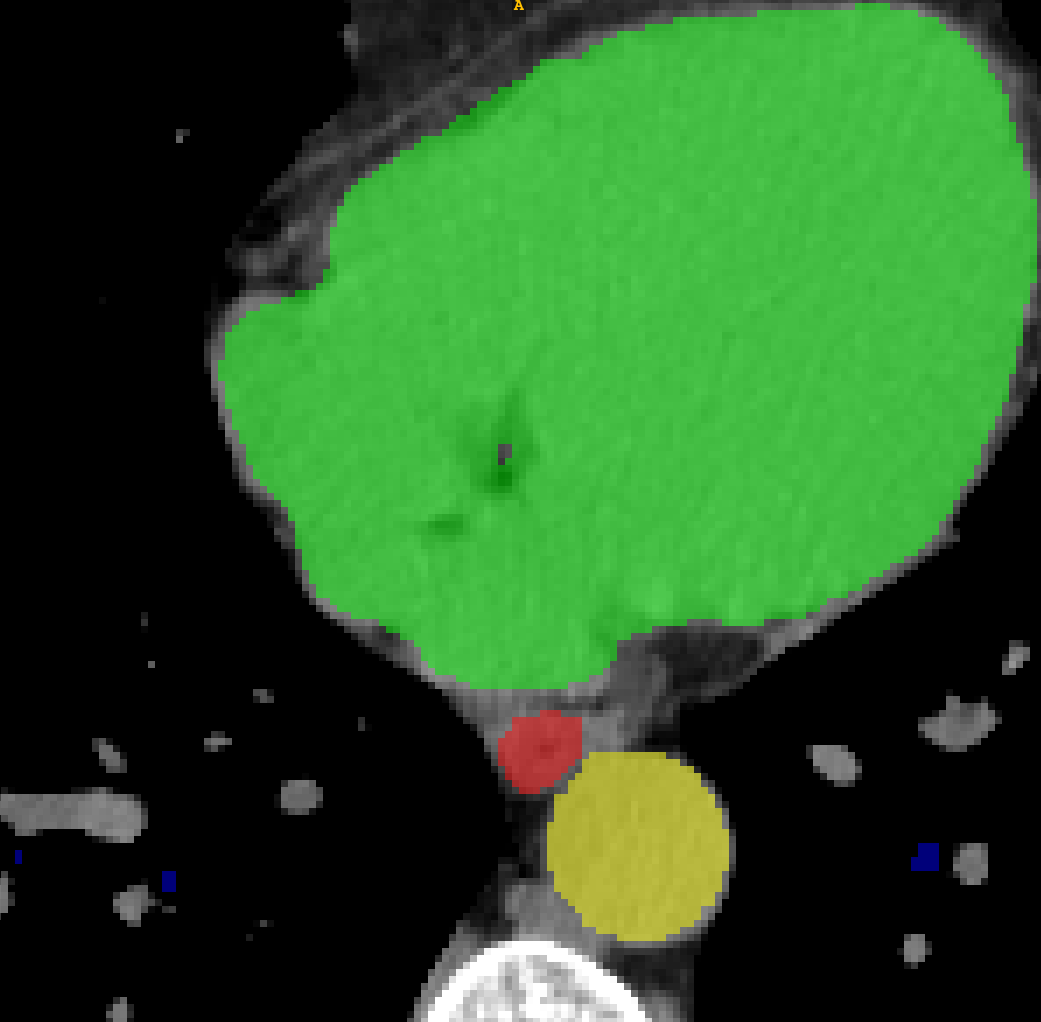}&
        \includegraphics[width=0.14\linewidth]{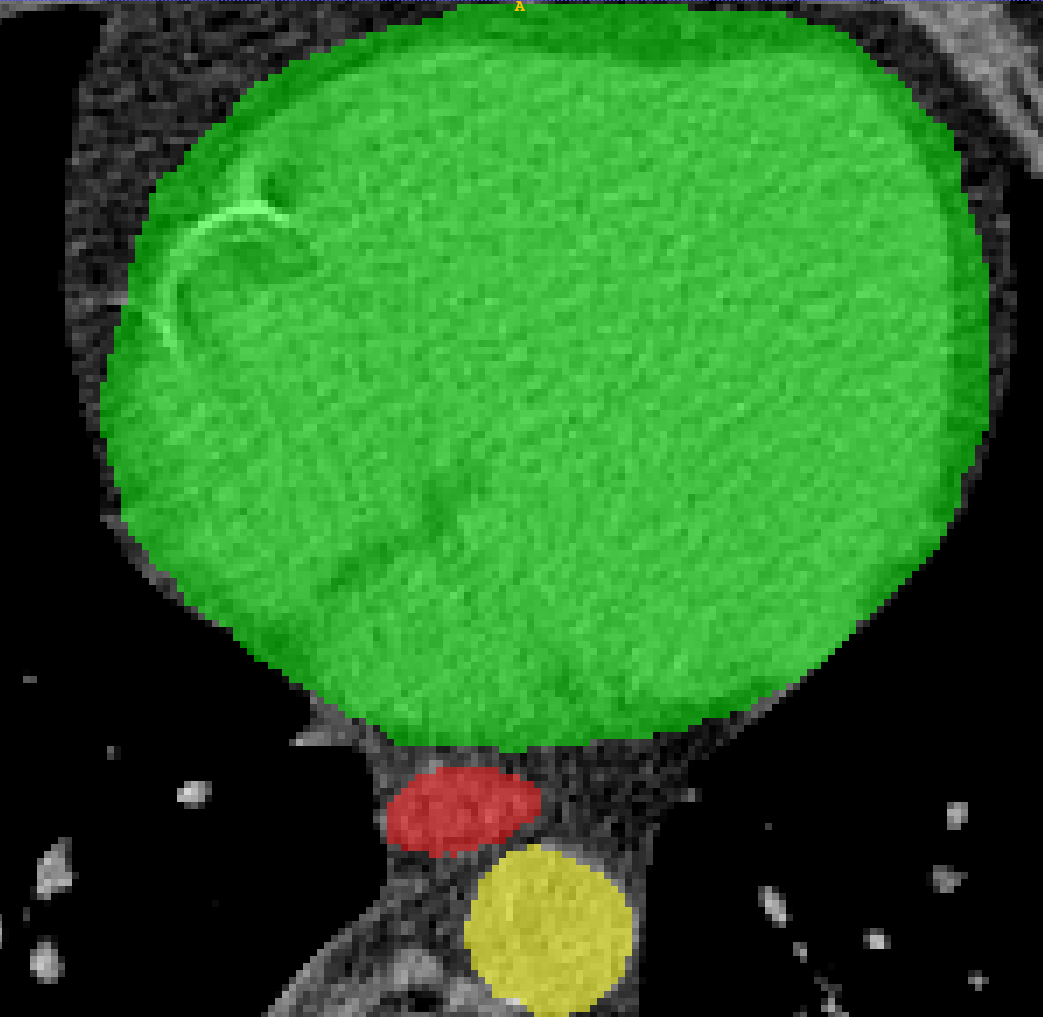}&
        \includegraphics[width=0.14\linewidth]{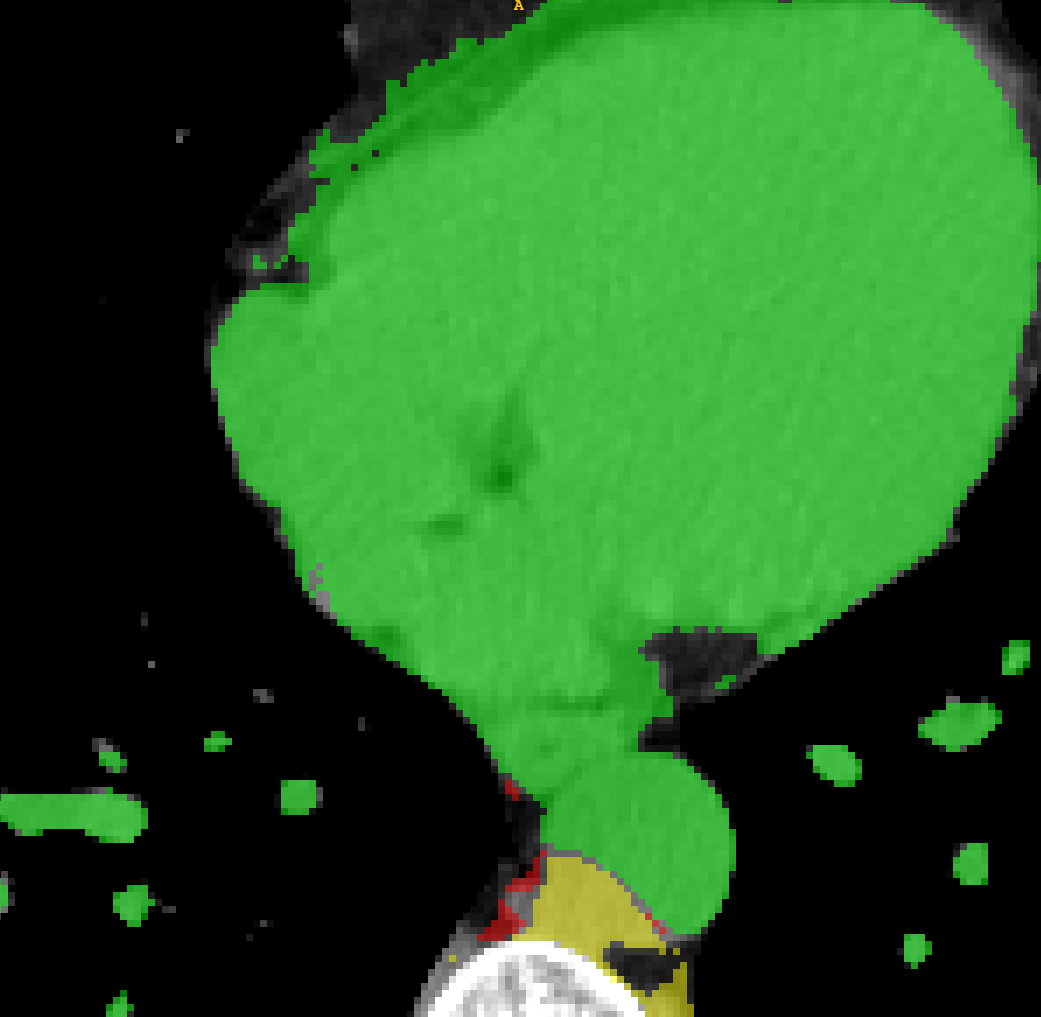}&
        \includegraphics[width=0.14\linewidth]{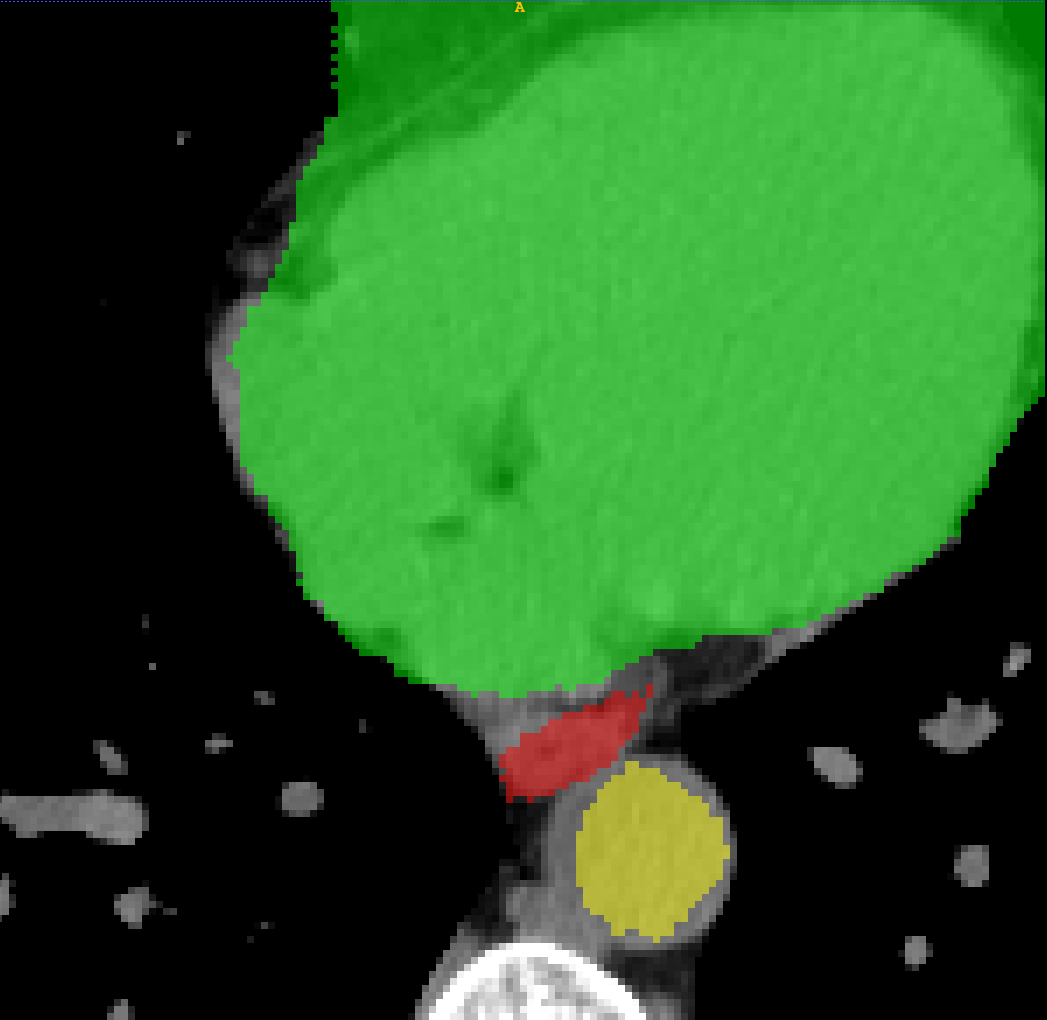}&
        \includegraphics[width=0.14\linewidth]{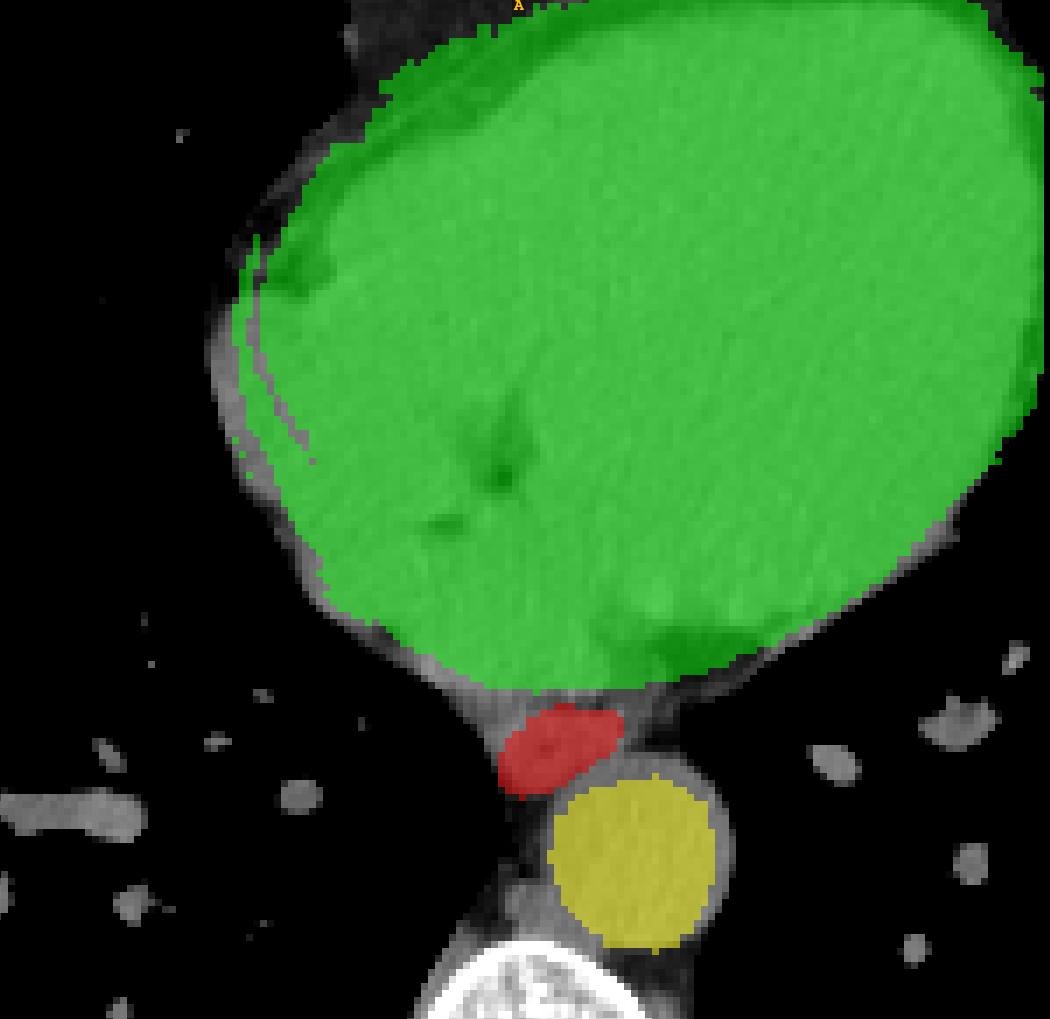}&
        \includegraphics[width=0.14\linewidth]{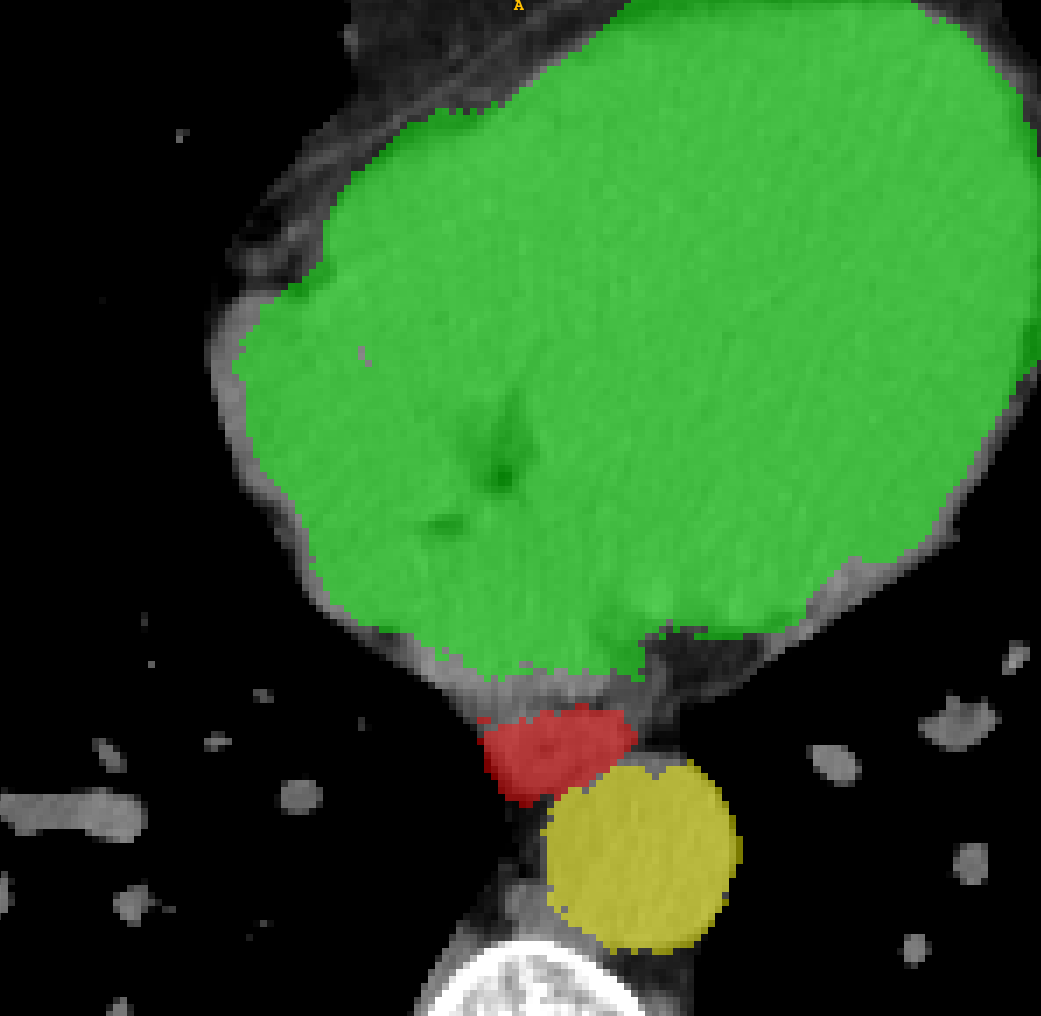}\\
        F& M& VoxelMorph& Ants& deedsBCV& Ours\\
    \end{tabular}
\end{table*}

\begin{table}[h]
    \centering
    \caption{Runtime of different methods. Ants and deedsBCV are open source C++ software package.}
    \label{tab:runtime}
    \begin{tabular}{lcc}
        \hline
        \ & GPU sec& CPU sec\\
        \hline
        $Ants\cite{Ants}$& -& 15.404\\
        $deedsBCV\cite{deedsBCV1,deedsBCV2}$& -& 17.390\\
        $RCN\cite{rcn}$& 0.786& -\\
        $VoxelMorph\cite{VoxelMorph4}$& 0.007& -\\
        $Ours$& 0.010& -\\
        \hline
    \end{tabular}
\end{table}

\subsection{Ablation Experiments}
\textbf{Refine Core}: RC is used to combine the information from current and previous stage.
If RC is removed, it is the same as the basic convolution block used in VoxelMorph.
Because the feature dimension is much less than VoxelMorph, which is only 3 in each up-sampling stage,
the dice score is lower than VoxelMorph, as shown in Table \ref{tab:ablation}.

\textbf{Rigid block}: We want to know if the rigid block is necessary.
So we designed the ablation experiment and the result is shown in Table \ref{tab:ablation}.
The rigid block can significantly enhance the effect of our registration algorithm.

\textbf{Regularization}: We verify the necessity of regularization loss function component $L_{range}$ and $L_{smooth}$, which is shown in Table \ref{tab:ablation}. In addition, we have done a lot of experiments using the grid search method to find the best combination of hyper-parameters $\alpha$ and $\beta$.
The best dice score is 68.42\% when $\alpha=10$ and $\beta=10^2$, which is shown in Figure \ref{fig:hyper-parameter}.

\subsection{Visualization}
We selected four images of large deformations occurring at different locations from different CT,
the visualization results are shown in Table \ref{tab:visualize}.

Rows 1-2 shows the CT gray image registration results.
The task is to register $M$ to $F$.
From the visualization of gray image results,
we can see that although ANTs and deedsBCV has high dice scores,
the visualization results look unrealistic.
Meanwhile, deep learning methods look smoother.
The sixth column shows the results of our method. 
The fitting effect is improved compared to previous methods.

Rows 3-4 shows the mask registration results.
The background of each image is $F$.
We could judge the performance by observing the fitting degree of the mask and background.
In many practical tasks, we use a registration algorithm to register $M$\rq{mask} to get the mask of $F$.
There may be a situation where the gray image fitting is good, but the mask result is poor.
This may be because the continuity of the registration field is not good enough.

\section{Conclusion}
This paper presents a novel deep learning model which refines the registration field in different resolution,
and verifies the effectiveness of processing images with large scale deformation and complex background. However, our method still could not achieve satisfactory regional continuity and need to use a lot of data for improving generalization. Further research will be carried out in the future.

{\small
\bibliographystyle{IEEEtran}
\bibliography{egbib}
}

\end{document}